\documentclass{article}
\usepackage{spconf,amssymb,amsmath,graphicx,tabularx,svg}

\title{Beyond Pixel Fidelity: Minimizing Perceptual Distortion and Color Bias in Night Photography Rendering}
\name{Furkan Kınlı}
\address{Bahçeşehir University\\
    Department of Artificial Intelligence Engineering\\
    İstanbul, Türkiye}

\begin{document}
%
\maketitle
\begin{abstract}
Night Photography Rendering (NPR) poses a significant challenge due to the extreme contrast between dark and illuminated areas in scenes, stemming from concurrent capture of severely dark regions alongside intense point light sources. Existing methods, which are mainly tailored for fidelity metrics, reveal considerable perceptual gaps and often detract from visual quality. We introduce pHVI-ISPNet, a novel RAW-to-RGB framework built on the robust HVI color space. Our network integrates four distinct key refinements: RAW-domain feature processing and Wavelet-based feature propagation to mitigate high-frequency detail loss; sample-based dynamic loss coefficients to ensure stable learning across varying exposure levels; and loss term based on feature distributions to maintain rigorous color constancy. Evaluations on the dataset introduced in the NTIRE 2025 challenge on NPR confirm our approach achieves competitive fidelity while establishing new state-of-the-art results in both CIE2000 color difference and LPIPS. This validates our perceptually-driven design for high-quality nighttime imaging.
\end{abstract}

\begin{keywords}
Night photography rendering, image signal processor, color correction, perceptual similarity
\end{keywords}
\section{Introduction}
\label{sec:intro}

The transformation of raw sensor-driven data from resource-limited devices to high-fidelity RGB-based imagery, in particular for Night Photography Rendering (NPR), represents an ongoing, pivotal challenge in computational photography. Although closely related to low-light image enhancement, NPR has its own very specific and more complicated set of problems, its contexts are very high-dynamic range scenes, where heavily underexposed areas are combined with intense, bright ones lit by different light sources. Moreover, these environments are characterized by extensive sensor noise and the extreme challenge of multiple, spectrally diverse illuminants, which significantly complicates accurate color correction and tone mapping. In these challenging scenarios, state-of-the-art methods frequently fail to correlate reliably with human visual quality, despite achieving high scores on traditional distortion metrics (\textit{i.e.}, PSNR and SSIM).  This limitation is further verified by empirical data from the NTIRE 2025 Challenge in NPR~\cite{Ershov_2025_CVPR}, which confirms this bias in prioritization: the ranking disparity between the objective leader and the human-preference leader underscores the perceptual gap.

\begin{figure}[!t]
\begin{minipage}[t]{.32\linewidth}
  \centering
  \centerline{\includegraphics[width=2.82cm]{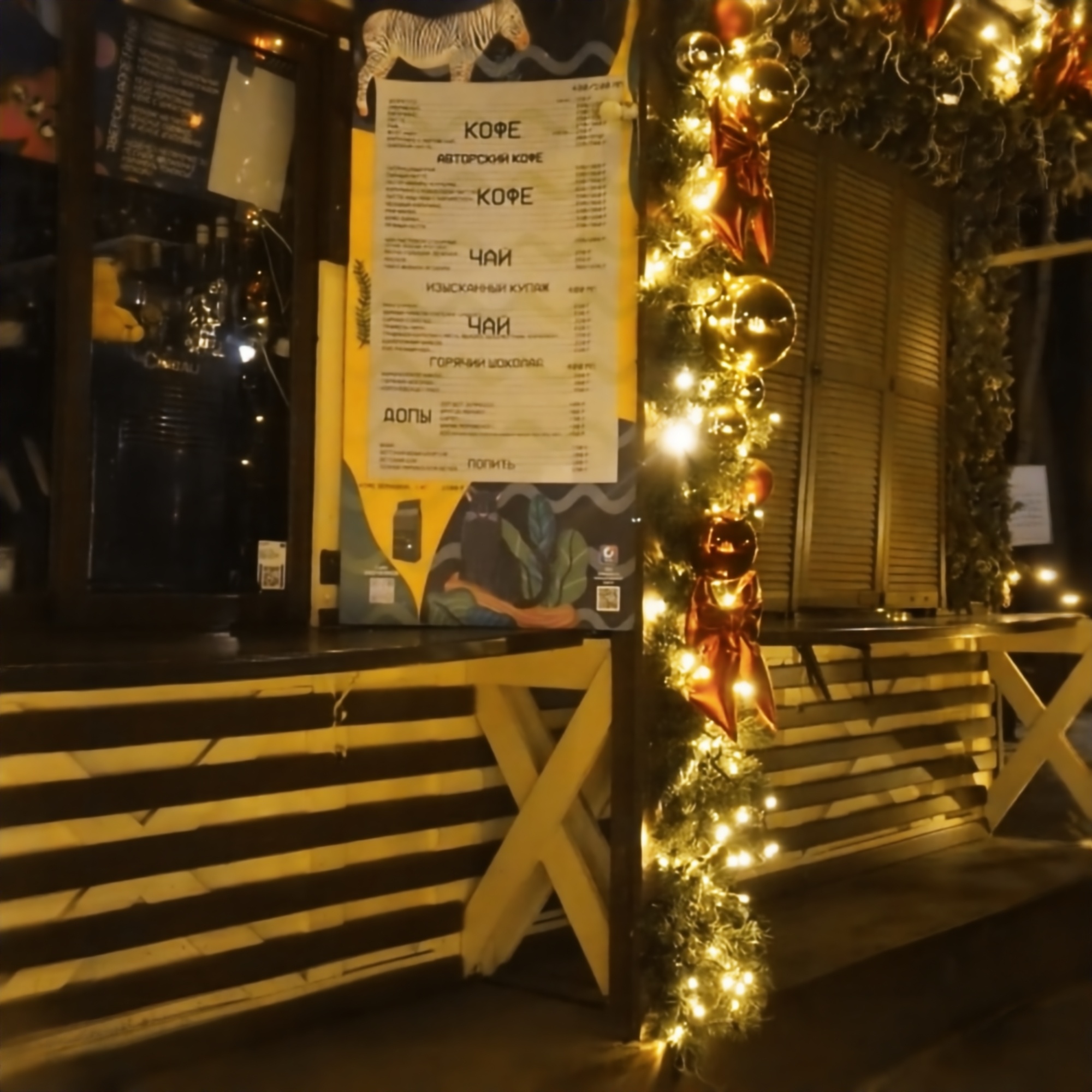}}
  \vspace{0.1cm}
  \centerline{\includegraphics[width=2.82cm]{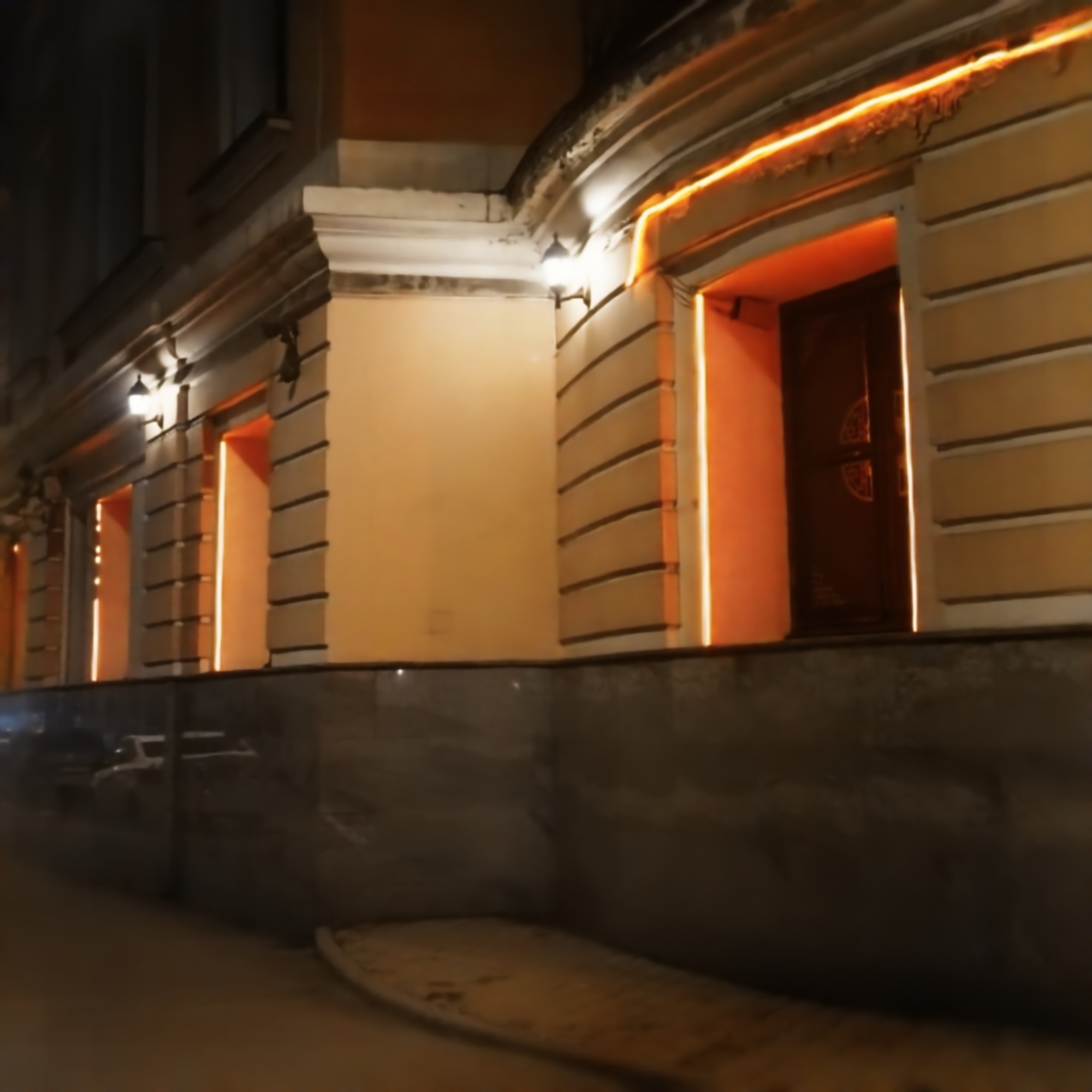}}
  \vspace{0.1cm}
  \centerline{\includegraphics[width=2.82cm]{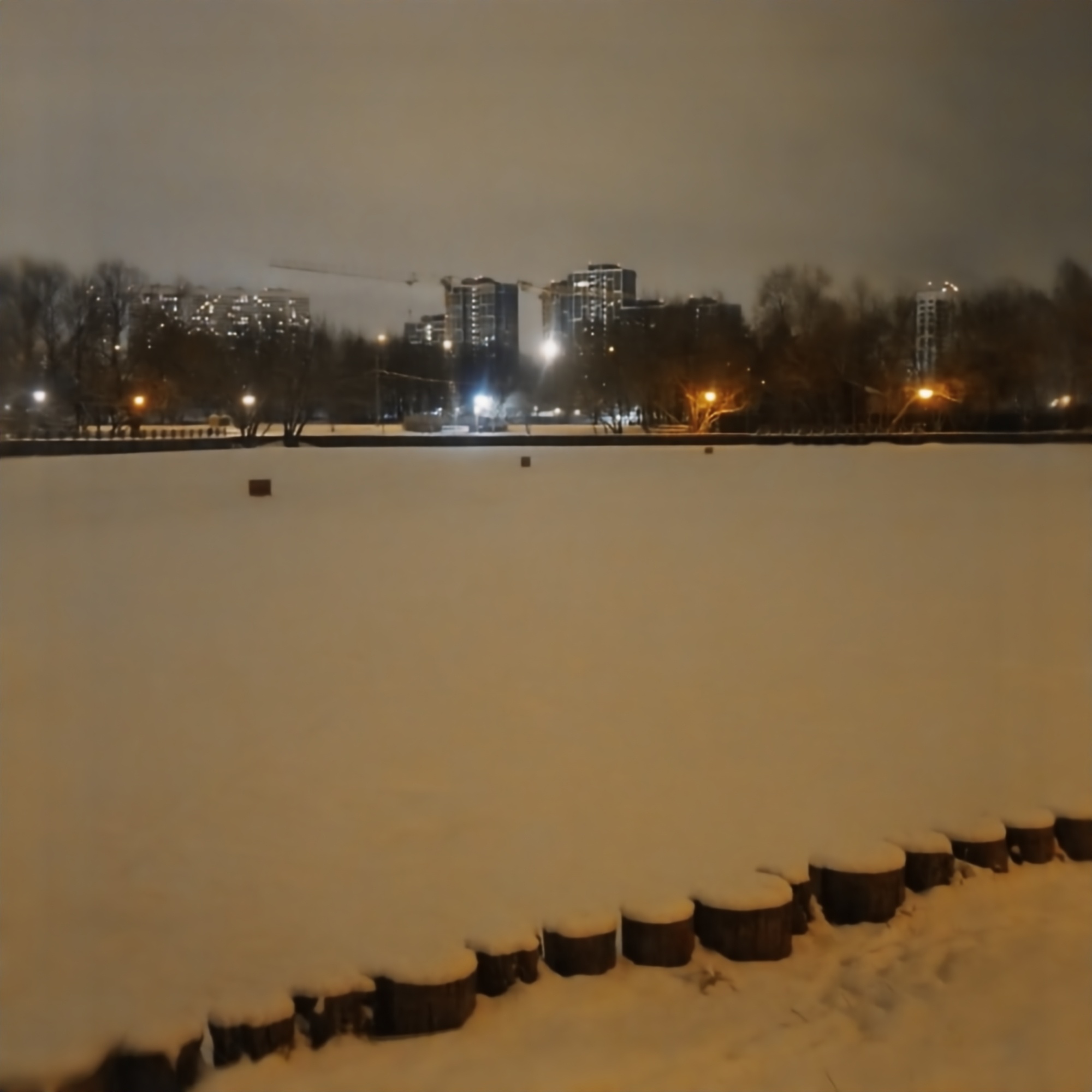}}
  \centerline{(a) NJUST\_KMG}\medskip
\end{minipage}
\hfill
\begin{minipage}[t]{.32\linewidth}
  \centering
  \centerline{\includegraphics[width=2.82cm]{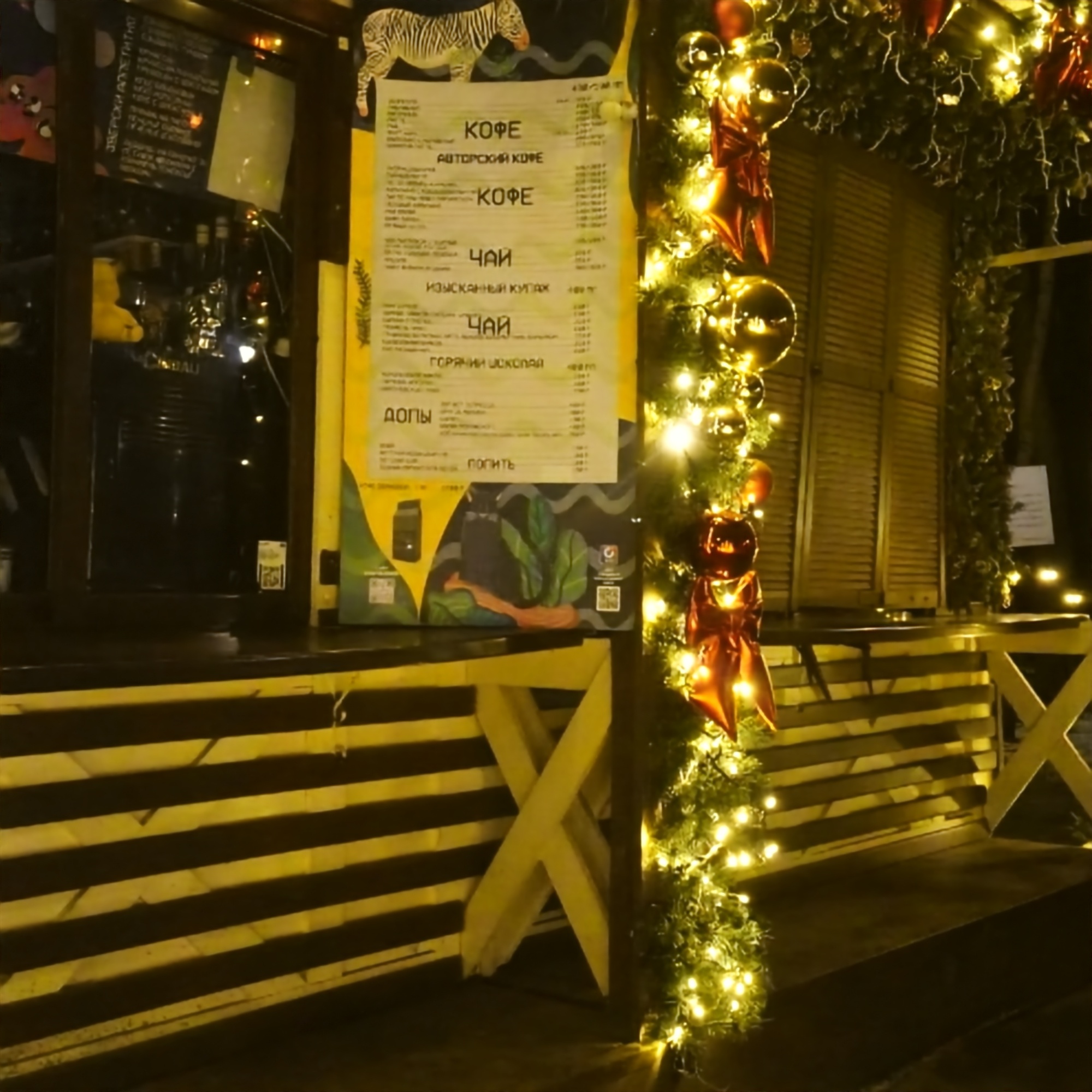}}
  \vspace{0.1cm}
  \centerline{\includegraphics[width=2.82cm]{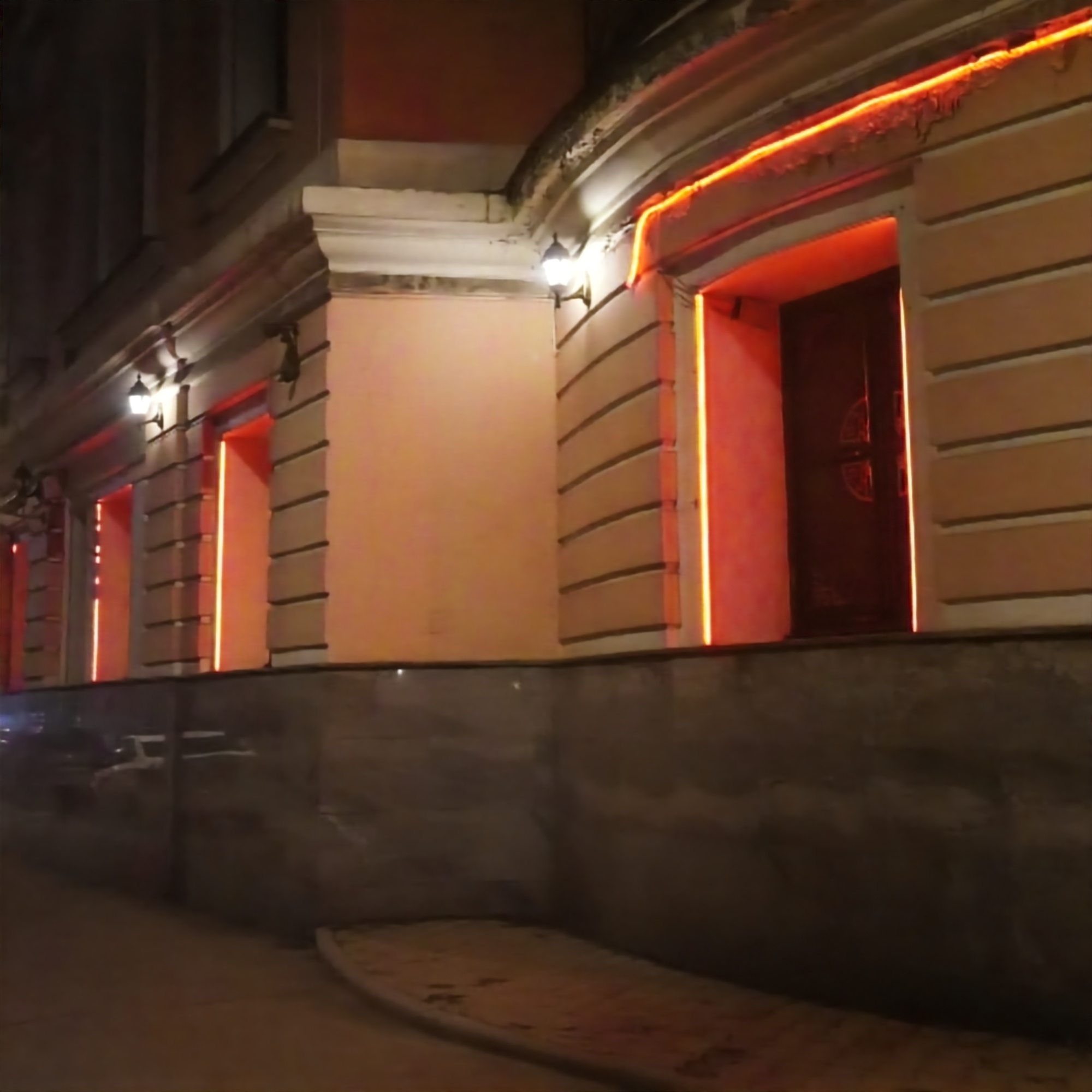}}
  \vspace{0.1cm}
  \centerline{\includegraphics[width=2.82cm]{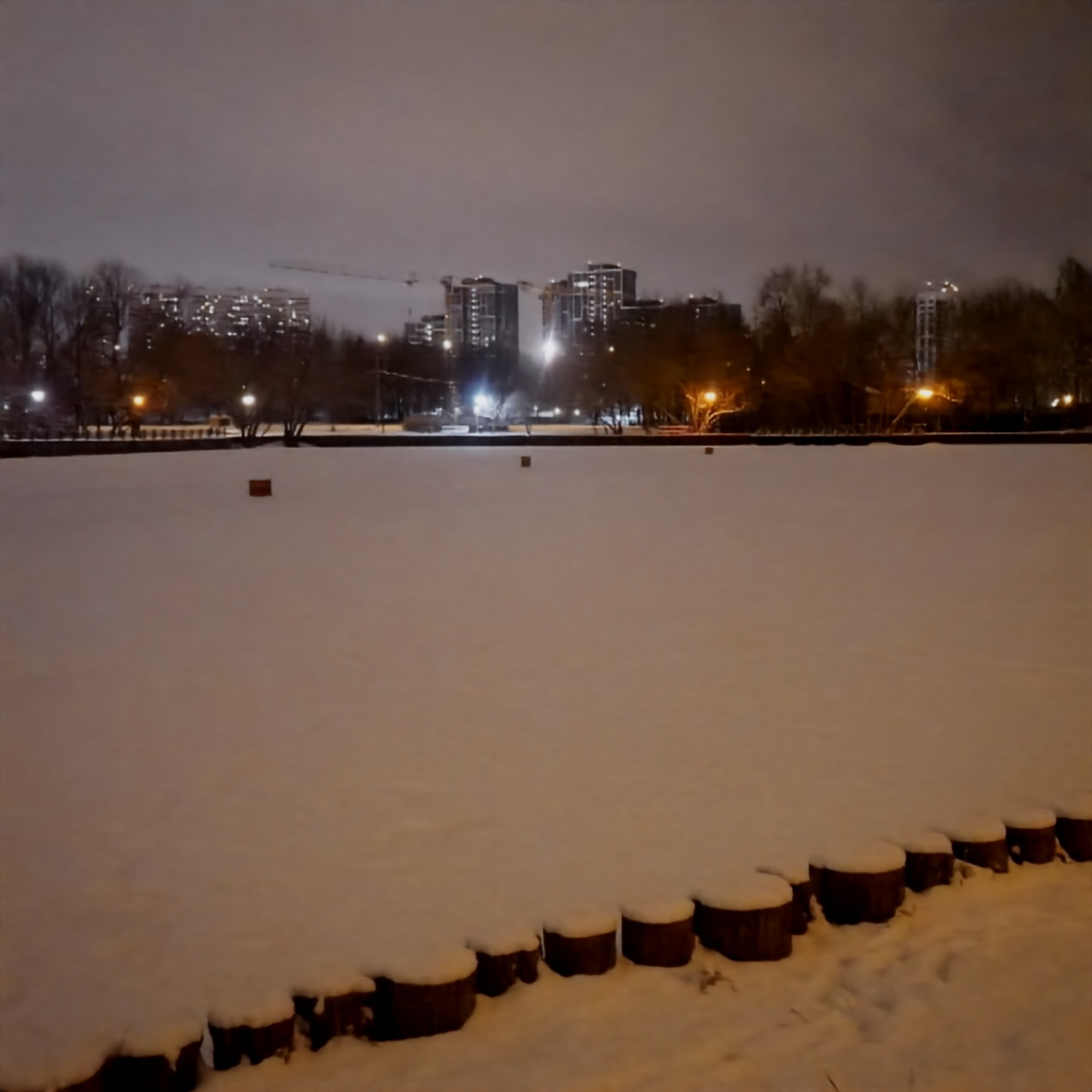}}
  \centerline{(b) Ours}\medskip
\end{minipage}
\hfill
\begin{minipage}[t]{0.32\linewidth}
  \centering
  \centerline{\includegraphics[width=2.82cm]{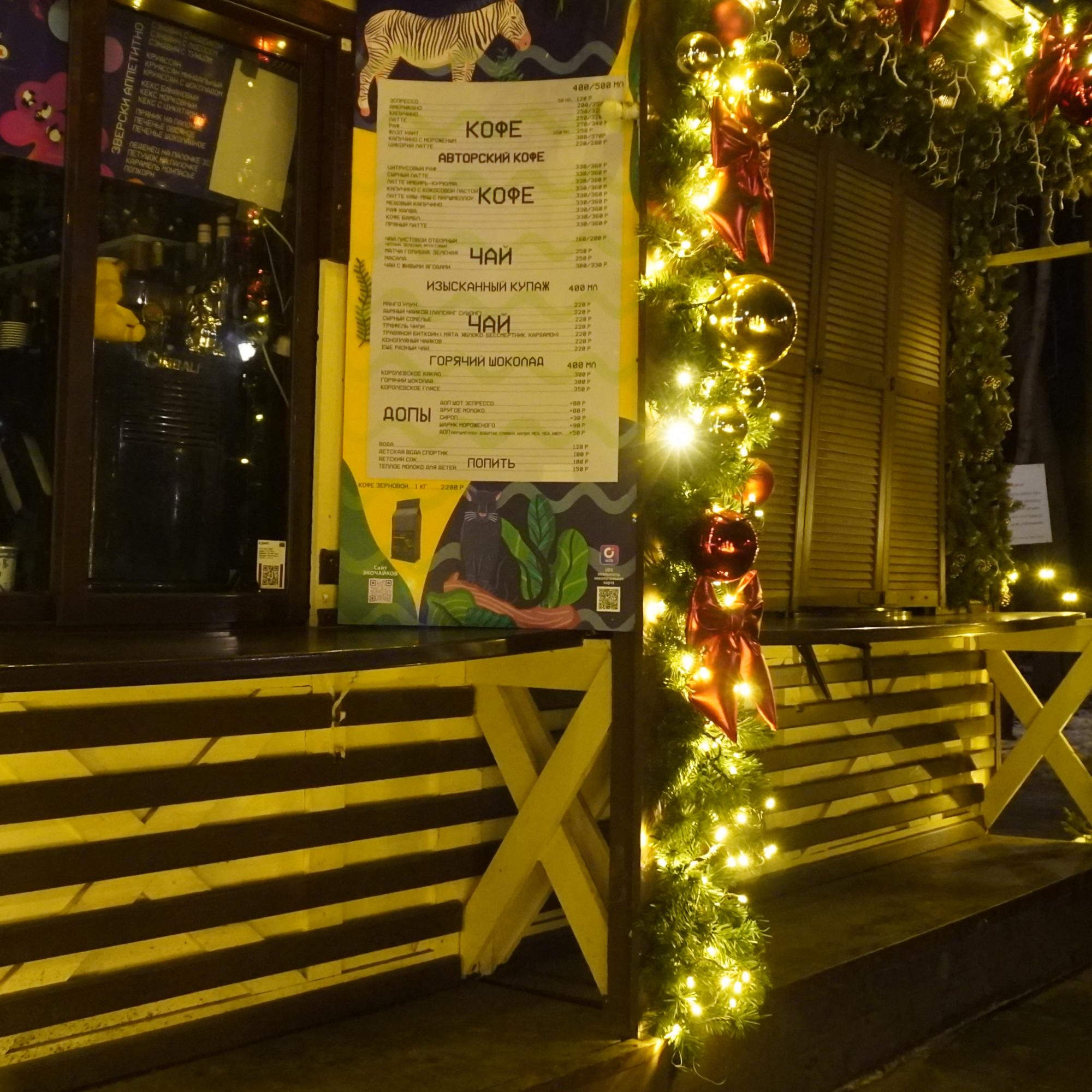}}
  \vspace{0.1cm}
  \centerline{\includegraphics[width=2.82cm]{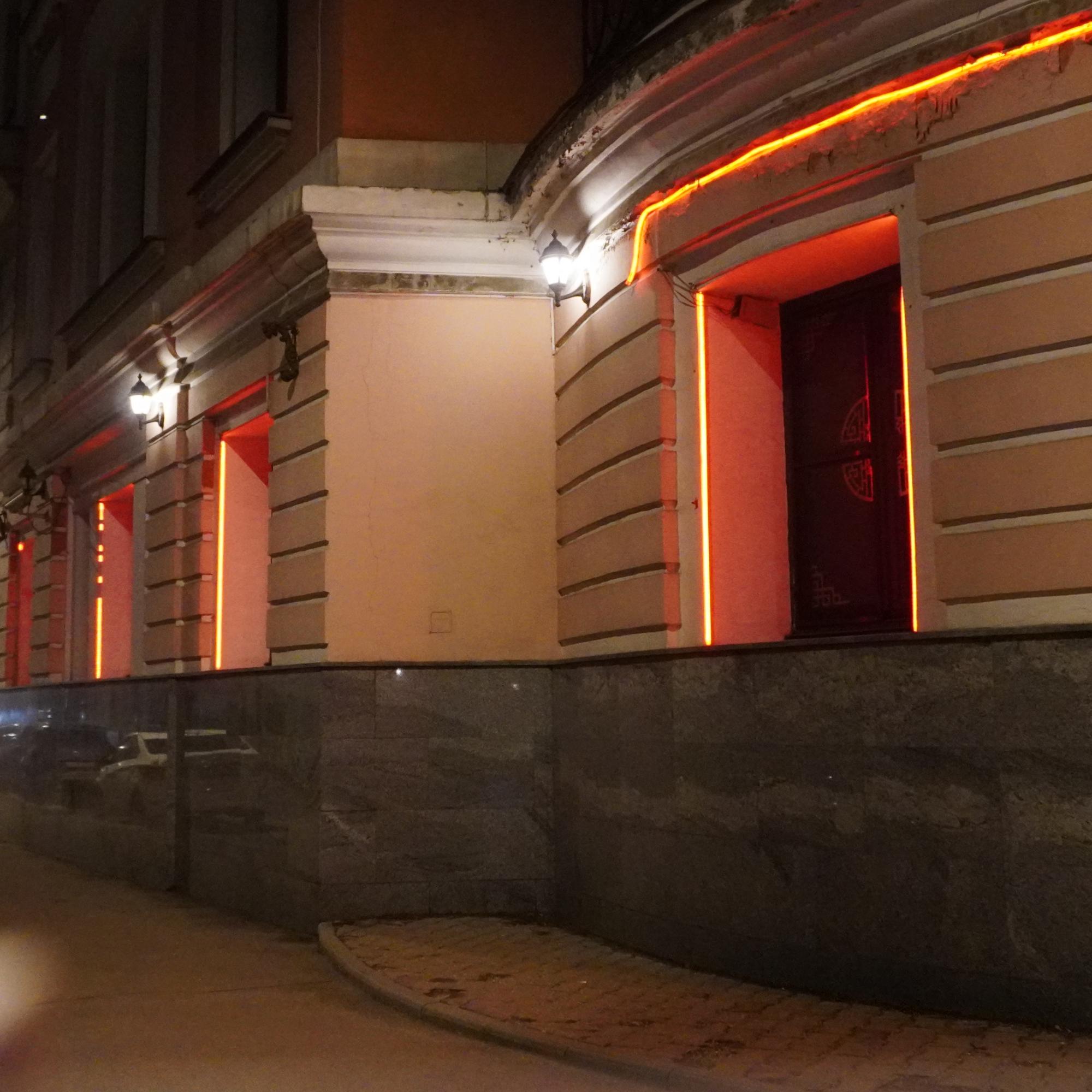}}
  \vspace{0.1cm}
  \centerline{\includegraphics[width=2.82cm]{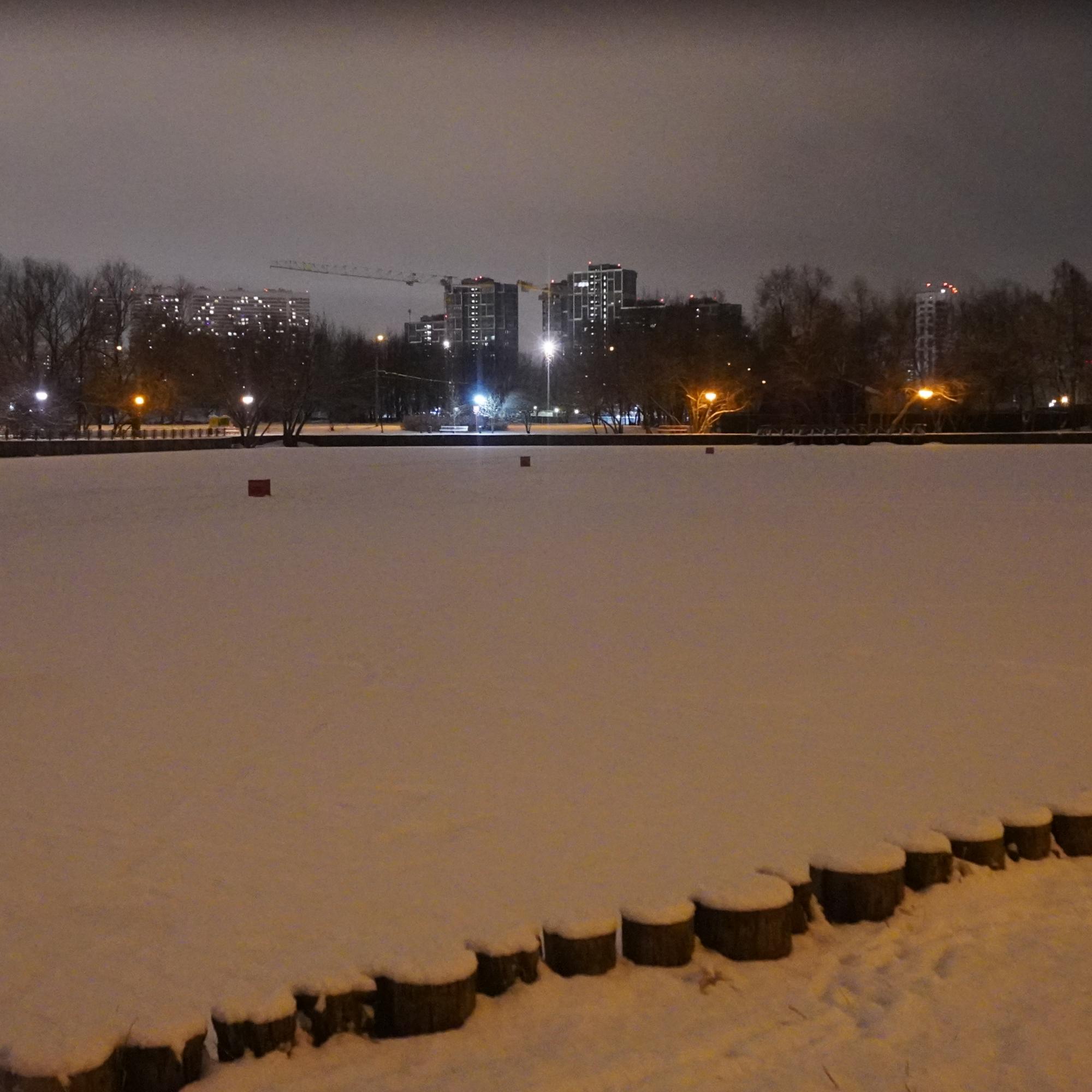}}
  \centerline{(c) GT}\medskip
\end{minipage}
\caption{Perceptual gap in challenging night-time scenes. While SOTA method (a) achieves high PSNR/SSIM, it suffers from severe perceptual distortion (\textit{e.g.}, color bias, tone artifacts) relative to GT (c). Our approach (b) delivers superior perceptual realism (\textit{i.e.}, faithful color, robust detail).}
\label{fig:perc_gap}
\end{figure}

To explicitly address this perceptual gap, we introduce \textit{pHVI-ISPNet}, a novel neural architecture for NPR. The network is built upon a color and intensity decoupling foundation, which utilizes the perceptually-motivated Horizontal/Vertical-Intensity (HVI) color space~\cite{yan2025hvi}. The HVI representation is strategically chosen because its foundation involves polarized Hue/Saturation (HS) maps and a learnable intensity collapse function. These features inherently mitigate the color bias and noise artifacts that are amplified when processing the complex and multi-illuminant night-time data in RGB or traditional HSV spaces.  This underlying structure, depicted in Figure~\ref{fig:perc_gap}, allows our network to bypass the severe color and tone compression artifacts that are a frequent issue with traditional reconstruction methods.

\textit{pHVI-ISPNet} significantly improves this core CIDNet architecture~\cite{yan2025hvi} with four domain-aware contributions to optimize the overall RAW-to-RGB rendering pipeline. We deploy an initial \textit{RGGB-to-Features} conversion block that directly receives the 4-channel RAW Bayer pattern input to give the network an opportunity to learn a feature-level demosaicking and noise management strategy at the sensor level. In addition, to minimize irreversible high-frequency textural degradation from traditional up/downsampling, we replace pooling and strided convolutions with wavelet-based transformations, as discussed in Liu \textit{et al.}~\cite{liu2018multi}. Our training is guided by two additional loss components: we incorporate FDM Loss~\cite{kinli2025feature}, based on Exact Feature Distribution Matching~\cite{zhang2022exact}, to minimize the output/ground truth feature distribution discrepancy. Finally, a new dynamic loss weighting strategy applies a sample-based coefficient to RGB and HVI loss terms, which calibrates their impact via the ratio of image mean intensities. Consequently, these refinements allow $\textit{pHVI-ISPNet}$ to surpass current limitations and achieve competitive fidelity while establishing new state-of-the-art results in both the CIE2000 color difference ($\Delta E$) and LPIPS.


\section{Related Works}
\label{sec:related}

Night Photography Rendering (NPR) is a challenging domain related to low-light image enhancement (LLIE), but distinct due to the necessity to process raw sensor data challenged by extreme noise, high-dynamic range conditions, and multi-illuminant color casts. The foundational LLIE work was based on the Retinex theory~\cite{land1971retinex}, which distinguishes illumination and reflectance, stimulating network designs such as RetinexNet~\cite{wei2018deep}. More recent LLIE methods focus on certain aspects of degradation. For instance, Bread~\cite{guo2023low} applies luminance-chrominance decomposition for noise and color distortion control, whereas generative methods such as Diff-Retinex~\cite{yi2023diff} build a new approach for enhancement by a Retinex decomposition coupled with a conditional diffusion model to recover missing information.

The field of NPR has been shaped by a series of annual workshops~\cite{Ershov_2022_CVPR,Shutova_2023_CVPR,banic2024ntire}, culminating in the NTIRE 2025 Challenge~\cite{Ershov_2025_CVPR}. This ongoing effort has led to the emergence of a major trend towards sophisticated, multi-stage pipelines that explicitly decouple denoising, demosaicking, and tone mapping. Key high-performance entries in this challenge included a dual attention-based architecture, Deep-FlexISP~\cite{liu2022deep} and HVI-CIDNet~\cite{yan2025hvi}. Crucially, the highest ranking objective solution (\textit{i.e.}, NJUST\_KMG) used a fusion strategy by which the output of the DNF (Decouple and Feedback) framework~\cite{jin2023dnf} was combined with that of MW-ISPNet~\cite{zhang2018image,liu2018multi}.


The color space selection is crucial for producing high-quality processing in extreme lighting. The standard RGB space has closely coupled brightness and color characteristics, severely amplifying noise and causing distortion. Conversely, while transforming to decoupled spaces such as HSV assists with overall tone control, the process introduces major, well-known types of artifacts in underexposed regions (\textit{i.e.}, red discontinuity noise and black plane noise). To overcome these fundamental limitations for low-light enhancement, Yan et al.~\cite{yan2025hvi} proposed Horizontal/Vertical-Intensity as the color space. HVI employs polarized Hue/Saturation maps and a learnable intensity collapse function, offering a robust mechanism for superior color and intensity decoupling. This effective color-handling mechanism is crucial for mitigating multi-illuminant color bias in NPR, and forms the main architectural basis of our proposed design.

\section{Methodology}
\label{sec:method}

\begin{figure*}[!t]
    \centering
    \centerline{\includesvg[width=\linewidth]{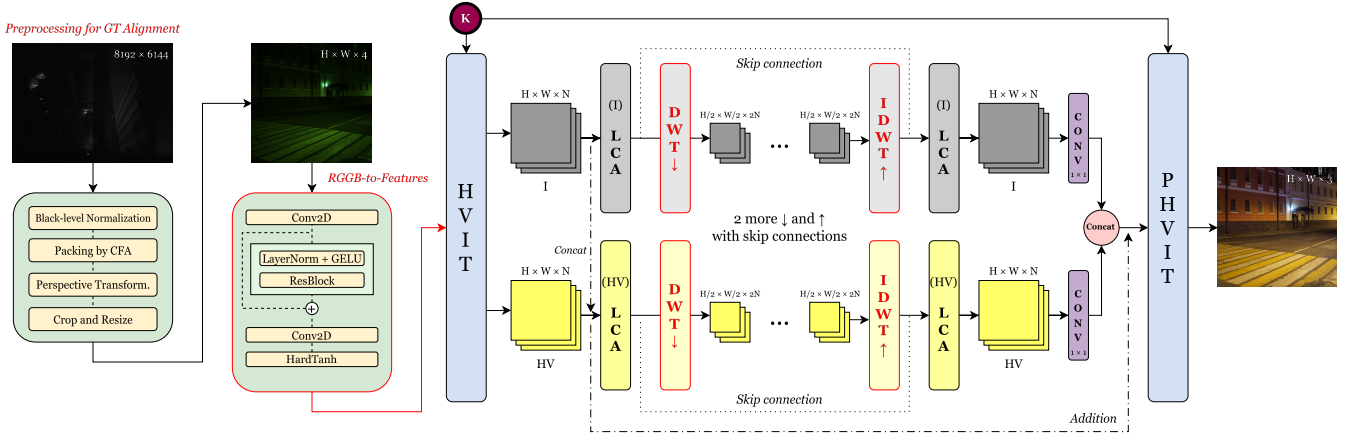}}

\caption{The pHVI-ISPNet Architecture. Our network extends the base CIDNet~\cite{yan2025hvi} dual-branch structure with two key structural innovations designed for NPR (\textit{i.e.}, the RAW-domain feature processing block and Wavelet-based feature propagation).}
\label{fig:arch}
\end{figure*}

The proposed \textit{pHVI-ISPNet} is a highly specialized architecture built on the Color and Intensity Decoupling Network (CIDNet)~\cite{yan2025hvi}, which utilizes the HVI color space to address the extreme high-dynamic range and multi-illuminant challenges in Night Photography Rendering (NPR). Our methodology incorporates four key innovations into the architectural and training pipeline, designed to maximize perceptual and color fidelity, as conceptually illustrated in Figure~\ref{fig:arch}.

\subsection{Base Architecture: CIDNet}

Our proposed network is built on the Color and Intensity Decoupling Network (CIDNet), which utilizes the HVI color space to mitigate severe color and noise artifacts. The network's core functionality centers on color and intensity decoupling: the HVI transformation ($T_{HVI}$) converts input features into complementary \textit{HV} features (\textit{i.e.}, color) and \textit{I} features (\textit{i.e.}, luminance). These streams are independently processed within a dual-branch U-Net structure. The aim of that separation is to minimize the color coupling and noise amplification inherent to RGB/HSV processing.

\subsection{RAW-Domain Feature Processing}

Unlike the base CIDNet, which accepts 3-channel RGB input, a sequence of initial operations is applied to the raw input data to ensure proper alignment and scaling. The RAW Bayer data undergoes standard preprocessing for ground truth alignment before feature extraction. This includes black/white level correction and subsequent packing into the 4-channel RGGB format $\mathbf{I}_{RAW} \in \mathbb{R}^{H \times W \times 4}$. This packed 4-channel data is then subjected to perspective transformation and cropping by given bounds to ensure precise pixel-level alignment with the ground truth image. 

The dedicated block $\textit{RGGB-to-Features}$ in our proposed architecture accepts the 4-channel input directly rather than the 3-channel input layer. This module consists of learned convolutional transformations that model the crucial initial RAW-domain operations (\textit{i.e.}, adaptive demosaicking and sensor-noise suppression). By taking this step within the feature domain, we ensure that the high-quality input features given to the HVI transformation (HVIT) are free of sensor-specific artifacts, allowing us to concentrate exclusively on specialized tone and color mapping.

\subsection{Wavelet-based Feature Propagation}

To prevent the irreversible degradation of essential high-frequency textural detail in the multi-scale feature compression of CIDNet, we employ Wavelet-based feature propagation. In the encoder, the downsampling is based on the DWT that leads to a near-lossless decomposition of the feature map into four sub-bands. Most importantly, instead of discarding the high-frequency sub-bands, all four of them are concatenated and fused with each other in a $1\times 1$ convolution. The decoder is symmetrically made to reconstruct features using the Inverse DWT (IDWT): the low-resolution features are channel-adjusted, split into low-pass and high-pass components, upsampled by the IDWT, and then concatenated into the skip connection for final feature fusion. This approach, inspired by~\cite{Ershov_2025_CVPR}, guarantees the continuity of the textural and edge information throughout the depth.

\subsection{Loss Functions}

The total supervision signal for \textit{pHVI-ISPNet} is defined by a highly compositional loss function $\mathcal{L}_{\text{Total}}$, designed to enforce structural fidelity, perceptual accuracy, and statistical consistency across both the final RGB output and the internal HVI representation.

The total loss is structured into two primary components: the dynamically weighted reconstruction term $\mathcal{L}_{p}$ and the specialized perceptual distortion terms $\mathcal{L}_{\Delta E}$ and $\mathcal{L}_{\text{FDM}}$.
\begin{equation}
\mathcal{L}_{\text{Total}} = \mathcal{L}_{p} + \lambda_{\Delta E} \cdot \mathcal{L}_{\Delta E} + \lambda_{\text{FDM}} \cdot \mathcal{L}_{\text{FDM}}
\end{equation}

$\mathcal{L}_{p}$ refers to a fusion of six distinct fidelity and structural terms, which are calculated in the RGB domain $\mathbf{I}_{\text{RGB}}$ and in the HVI domain $\mathbf{I}_{\text{HVI}}$. All these six terms are scaled by a novel sample-based dynamic loss weighting scheme that provides stable training and ensures a balanced contribution of samples across the wide exposure range of scenes. This mechanism penalizes significant mean luminance deviation in both directions (\textit{i.e.}, too dark or too bright), as it requires the network to maintain a stable, balanced luminance ratio across the wide exposure range.

The coefficent $\boldsymbol{\alpha} \in \mathbb{R}^N$ is defined element-wise for a batch of $N$ samples where $\alpha_i$ is the coefficient for the $i$-th sample based on the ratio of global mean intensities ($\mu$).
\begin{equation}
\alpha_i = \max \left( \frac{\mu_{\text{GT}}^i}{\mu_{\text{pred}}^i}, \frac{\mu_{\text{pred}}^i}{\mu_{\text{GT}}^i} \right)
\end{equation}

The total reconstruction loss $\mathcal{L}_{p}$ across the batch is then calculated as the weighted average of the losses per-sample.
\begin{equation}
\mathcal{L}_{p} = \frac{1}{N} \sum_{i=1}^{N} \alpha_i \cdot \sum_{I \in \{\mathbf{I}_{\text{RGB}}, \mathbf{I}_{\text{HVI}}\}} \left( \mathcal{L}_{\text{L1}}^i(I) + \mathcal{L}_{\text{SSIM}}^i(I) + \mathcal{L}_{\text{E}}^i(I) \right)
\end{equation}

This procedure increases the loss for deviations from unity in either direction, prevents local minima from collapsing, and ensures luminance correction for a wide exposure range. Likewise, as mentioned in \cite{yan2025hvi}, $\mathcal{L}_{\text{L1}}$ is responsible for pixel fidelity, $\mathcal{L}_{\text{SSIM}}$ enforces structural similarity, and $\mathcal{L}_{\text{E}}$ retains important information about the high-frequency edge.

Furthermore, we employ specialized losses to maximize the impact of direct perceptual and statistical supervision on RGB output. First, we incorporate the CIE2000 color difference loss $\mathcal{L}_{\Delta E}$, which rigorously quantifies the color fidelity in a perceptually uniform color space. This loss provides a direct signal that correlates precisely with color accuracy. Another specialized objective that we use is Feature Distribution Matching (FDM) Loss ($\mathcal{L}_{\text{FDM}}$)~\cite{kinli2025feature}, derived from the Exact Feature Distribution Matching (EFDM)~\cite{zhang2022exact}. By utilizing EFDM to align the empirical feature distributions of the predicted and ground truth output, we enforce robust color and style consistency, directly leveraging its proven capability for managing multi-illuminant variations within the NPR domain. The FDM loss is mathematically defined as follows.
\begin{equation}
\mathcal{L}_{\text{FDM}}(f_{\text{pred}}, f_{\text{gt}}) = \ \frac{1}{n} \sum_{i=1}^{n}
\left[ f_{\text{pred}}[i] - f_{\text{gt}}[\text{rank}(f_{\text{pred}}[i])] \right]^2
\end{equation}
where $f_{\text{pred}}[i]$ and $f_{\text{gt}}[i]$ are elements of the predicted and ground-truth feature vectors (\textit{e.g.}, [CLS] tokens of ViT~\cite{dosovitskiy2021an}), respectively. The function $\text{rank}(\cdot)$ maps each element of $f_{\text{pred}}$ to its corresponding position in the sorted vector $f_{\text{gt}}$.

\section{Experiments}
\label{sec:experiments}
\subsection{Experimental Setup}
We perform all training and evaluation on the dataset given by the NTIRE 2025 Challenge on Night Photography Rendering (NPR). This dataset consists of paired low-light RAW images captured by a mobile phone (Huawei) and corresponding high-quality RGB images (Sony) serving as ground truth. For training, we utilize random cropping with a patch size of $768 \times 768$ pixels. Evaluation is performed on full-resolution $200$ images (\textit{i.e.}, $2000 \times 2000$) delivered for the final testing phase to ensure direct comparability with the studies outlined in this challenge.

\textit{pHVI-ISPNet} is implemented in PyTorch~\cite{paszke2019pytorch} and trained on 2 $\times$ NVIDIA RTX 2080Ti GPUs. The training consists of 2500 epochs with a batch size of 2. We utilize the AdamW optimizer~\cite{loshchilov2017decoupled} with an initial learning rate of $2 \times 10^{-4}$. The learning rate schedule incorporates a linear warmup over the first 3 epochs, followed by a cosine annealing schedule that gradually reduces the rate to a minimum of $1 \times 10^{-5}$.

We assess the performance using multiple metrics to validate both reconstruction fidelity and perceptual realism. We report conventional distortion metrics: PSNR and SSIM. Most importantly, our primary focus is on perceptual distortion and color accuracy metrics: LPIPS~\cite{zhang2018perceptual}, which quantifies the perceptual distance in the feature space, and the $\text{CIE}2000$ color difference ($\Delta E$)~\cite{luo2001development}. This focus is essential to demonstrate that our architecture alleviates the perceptual and color fidelity limitations of existing methods, which largely exploits pixel-wise reconstruction losses with maximum increase on PSNR/SSIM scores.

\begin{table}[!t]
    \caption{Quantitative Comparison on Night Photography Rendering benchmark~\cite{Ershov_2025_CVPR}.}
    \centering
    \label{tab:comp_results}
    \begin{tabularx}{\columnwidth}{l|c|c|c|c}
    \hline
    \textbf{Method} & \textbf{PSNR} $\uparrow$ & \textbf{SSIM} $\uparrow$ & $\mathbf{\Delta E} \downarrow$ & \textbf{LPIPS} $\downarrow$ \\
    \hline
    OzU-VGL & 18.93 & 0.677 & 11.94 & 0.586 \\
    colab & 19.15 & 0.635 & 10.27 & 0.476 \\
    AITH ITMO & 20.58 & 0.718 & --- & --- \\
    POLYU-AISP & 21.23 & 0.737 & 7.53 & 0.463 \\
    psykhexx & 21.70 & 0.754 & 7.48 & 0.455 \\
    DnF~\cite{jin2023dnf} & 23.06 & 0.781 & 6.41 & 0.454 \\
    NoTeam & 21.22 & 0.699 & 8.62 & 0.453 \\
    PSU & 22.21 & 0.758 & 7.04 & 0.441 \\
    NJUST\_KMG & \textbf{23.82} & \textbf{0.793} & 5.85 & 0.433 \\
    sorange & 21.16 & 0.744 & 7.82 & 0.429 \\
    Mialgo & 23.06 & 0.784 & 6.52 & 0.400 \\
    \textit{pHVI-ISPNet} & 23.63 & 0.785 & \textbf{5.42} & \textbf{0.388} \\
    \hline
    \end{tabularx}
\end{table}

\subsection{Comparative Results}

We analyze the performance of \textit{pHVI-ISPNet} against several state-of-the-art models on the NTIRE 2025 NPR Challenge benchmark~\cite{Ershov_2025_CVPR}, including top-performing approaches such as DnF~\cite{jin2023dnf} \& MW-ISPNet~\cite{liu2018multi} (NJUST\_KMG) and Deep-FlexISP-V2~\cite{liu2022deep} (Mialgo). The comparative results, averaged on the set given in the testing phase, are summarized in Table~\ref{tab:comp_results}, sorted by LPIPS to highlight perceptual performance.

The quantitative results in Table~\ref{tab:comp_results} confirm the motivation of this study: simply optimizing for the pixel fidelity metrics is insufficient for the perceptual realism in NPR. Although the fidelity leader achieves the highest PSNR (23.82) and SSIM (0.793), its performance measured using perceptual metrics ($\Delta E$: 5.85, LPIPS: 0.433) supports the long-standing trade-off between reconstruction accuracy and human visual quality. In contrast, \textit{pHVI-ISPNet} shows a paradigm shift by bridging this gap. Our approach achieves highly competitive fidelity metrics (PSNR: 23.63, SSIM: 0.785), which confirms robust reconstruction, but fundamentally advances the state-of-the-art in perceptual realism. Specifically, \textit{pHVI-ISPNet} establishes the lowest scores for both the CIE2000 color difference metric ($\Delta E$: $5.42$) and LPIPS ($0.388$). This result surpasses not only the fidelity leader, but also the previous best perceptual model, Mialgo ($\Delta E$: $6.52$, LPIPS: $0.400$).

\begin{figure*}[!t]
\begin{minipage}[t]{.16\linewidth}
  \centering
  \centerline{\includegraphics[width=2.82cm]{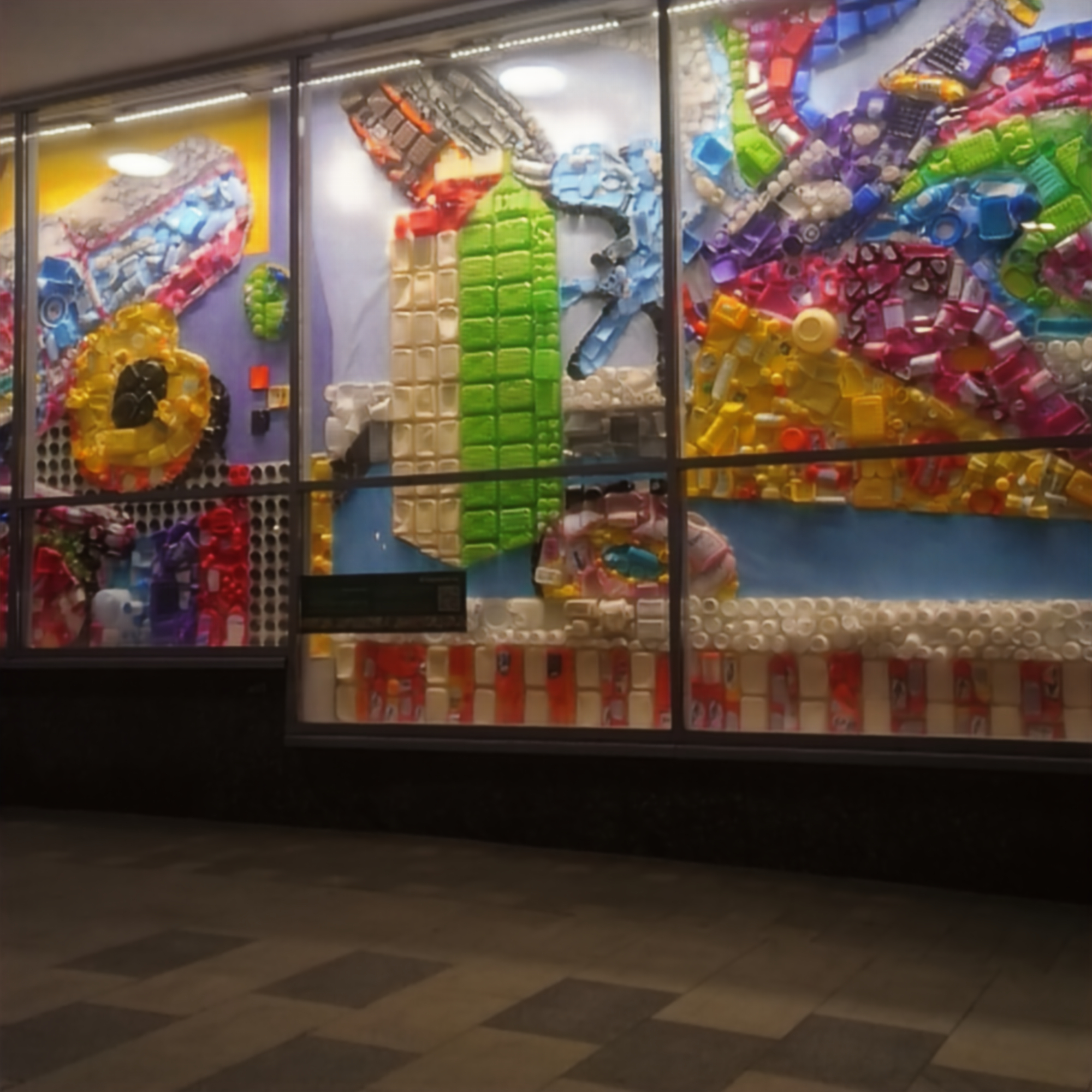}}
  \vspace{0.1cm}
  \centerline{\includegraphics[width=2.82cm]{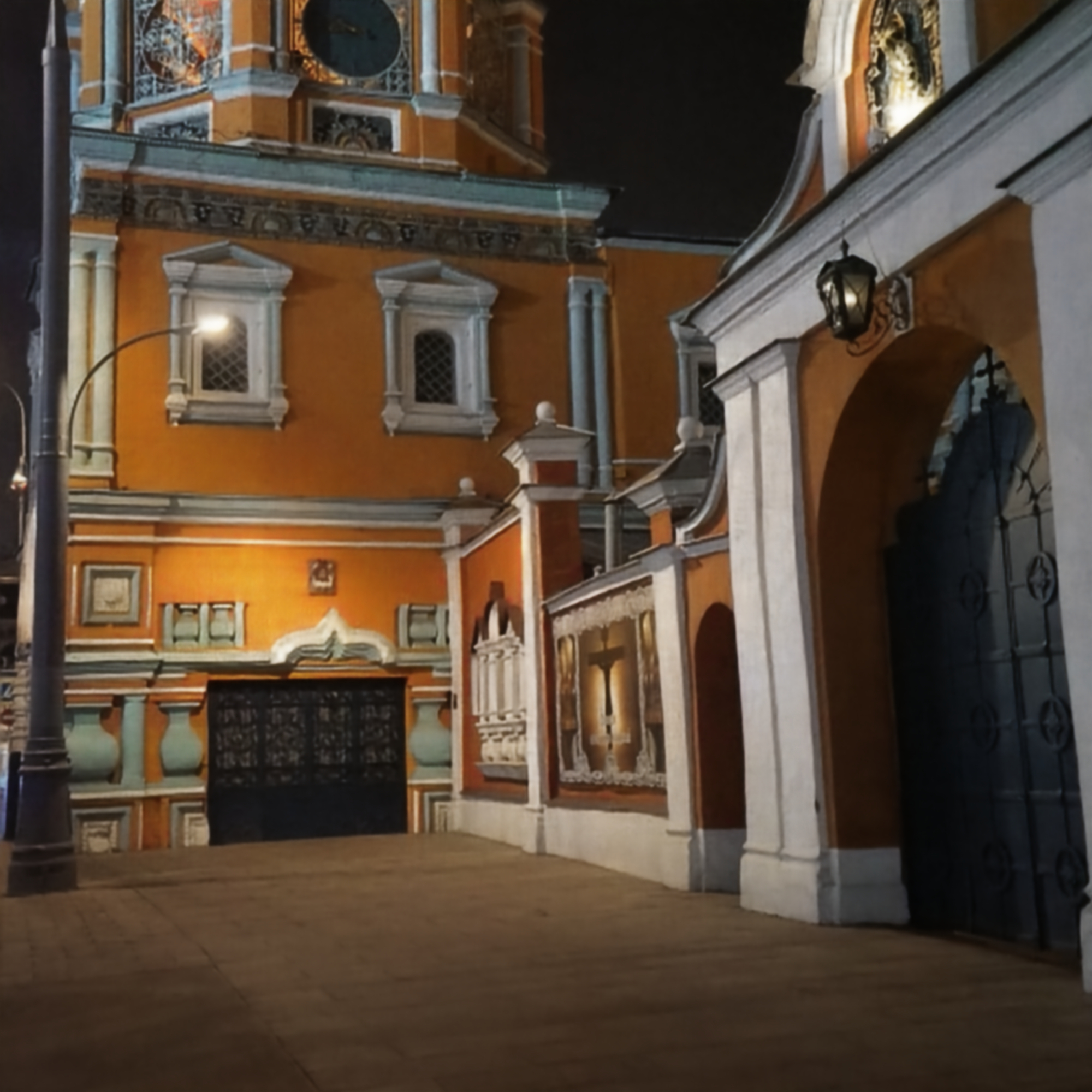}}
  \vspace{0.1cm}
  \centerline{\includegraphics[width=2.82cm]{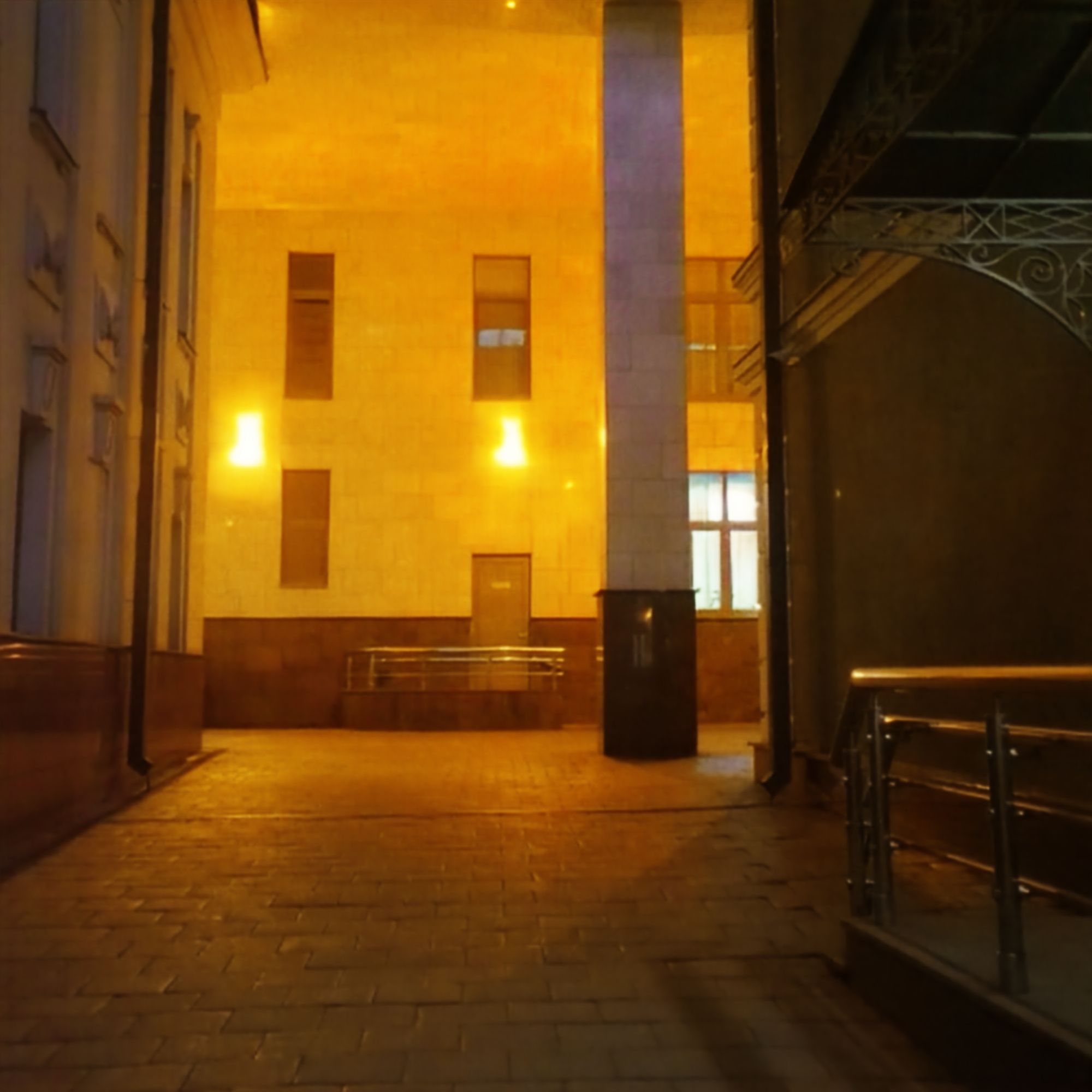}}
  \centerline{PSU}\medskip
\end{minipage}
\hfill
\begin{minipage}[t]{.16\linewidth}
  \centering
  \centerline{\includegraphics[width=2.82cm]{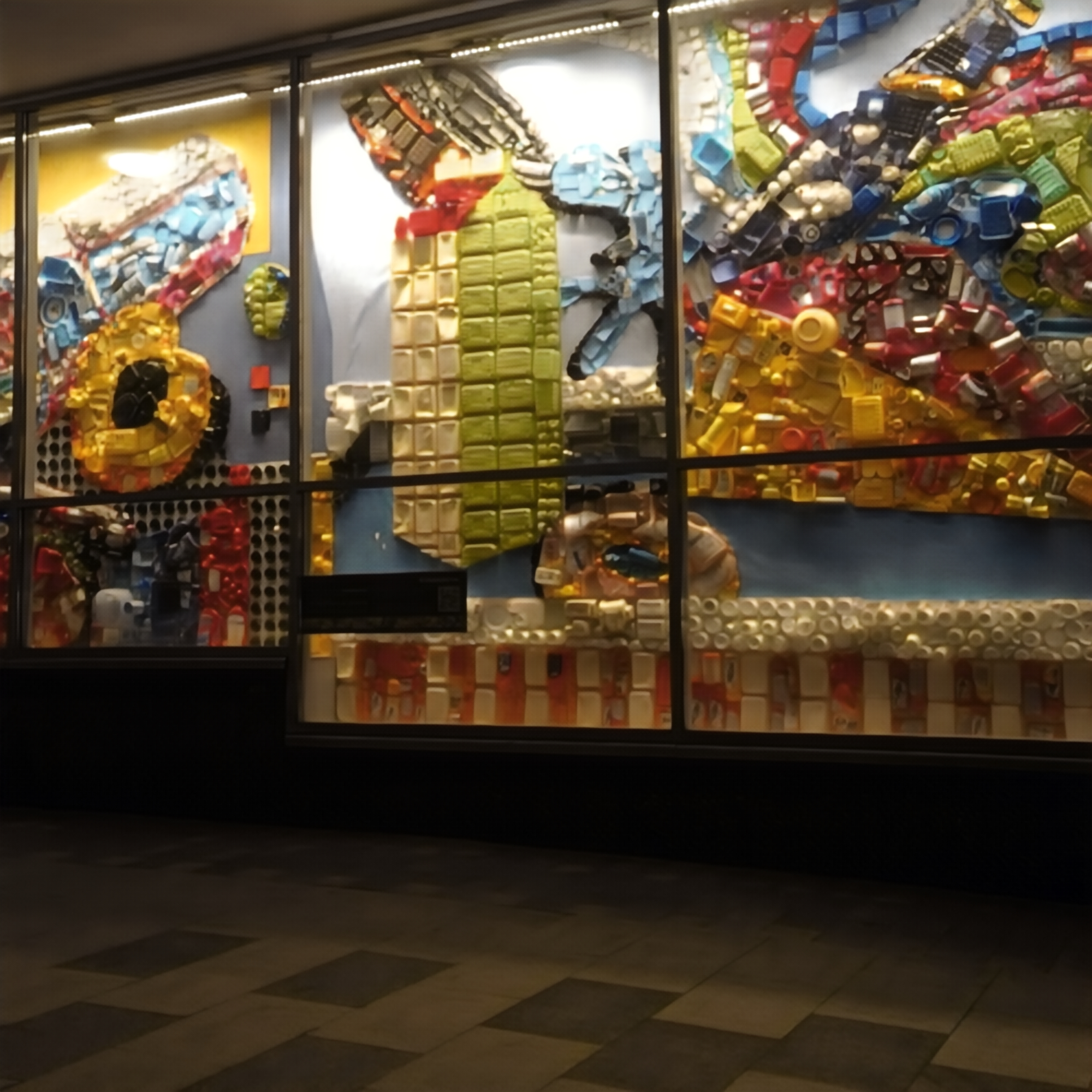}}
  \vspace{0.1cm}
  \centerline{\includegraphics[width=2.82cm]{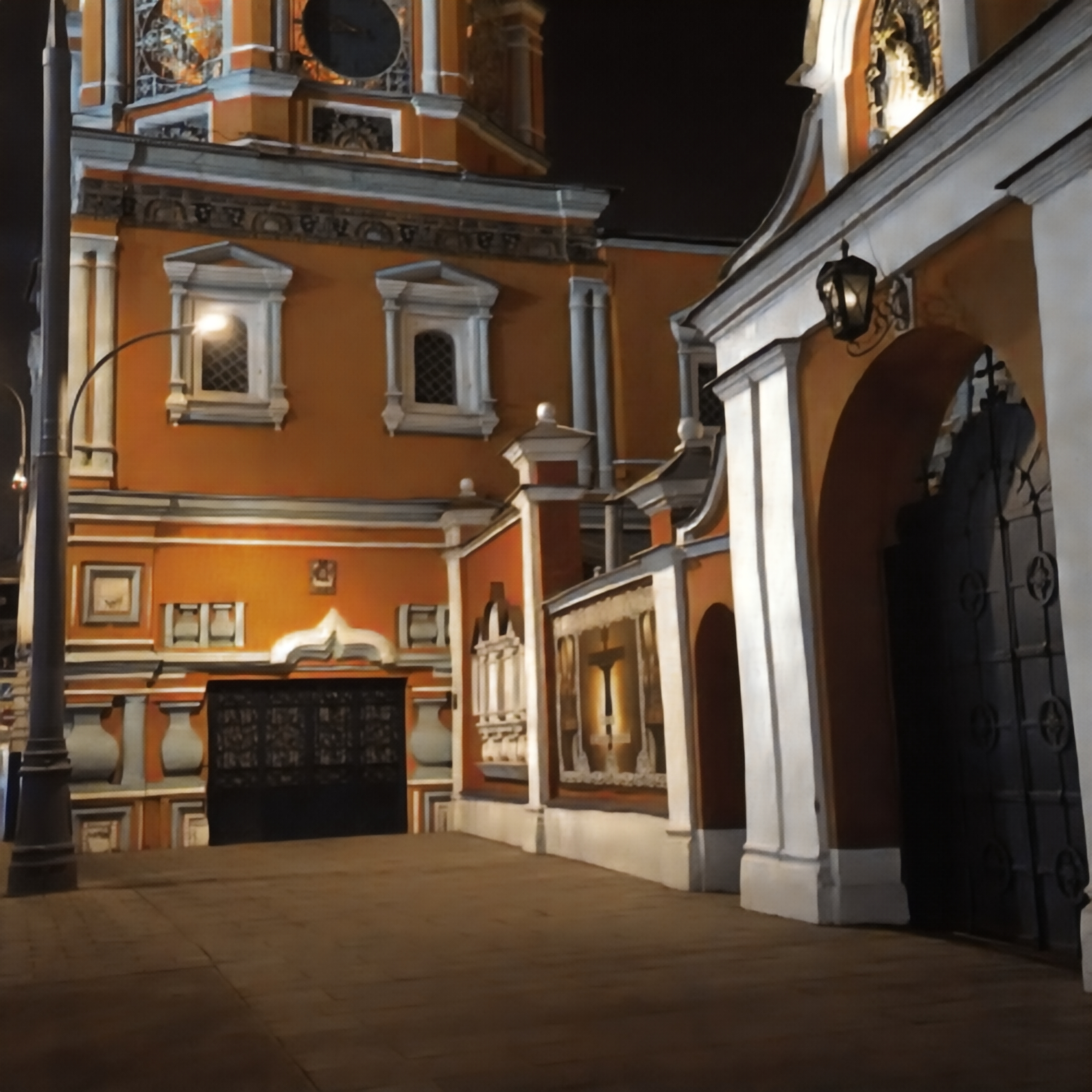}}
  \vspace{0.1cm}
  \centerline{\includegraphics[width=2.82cm]{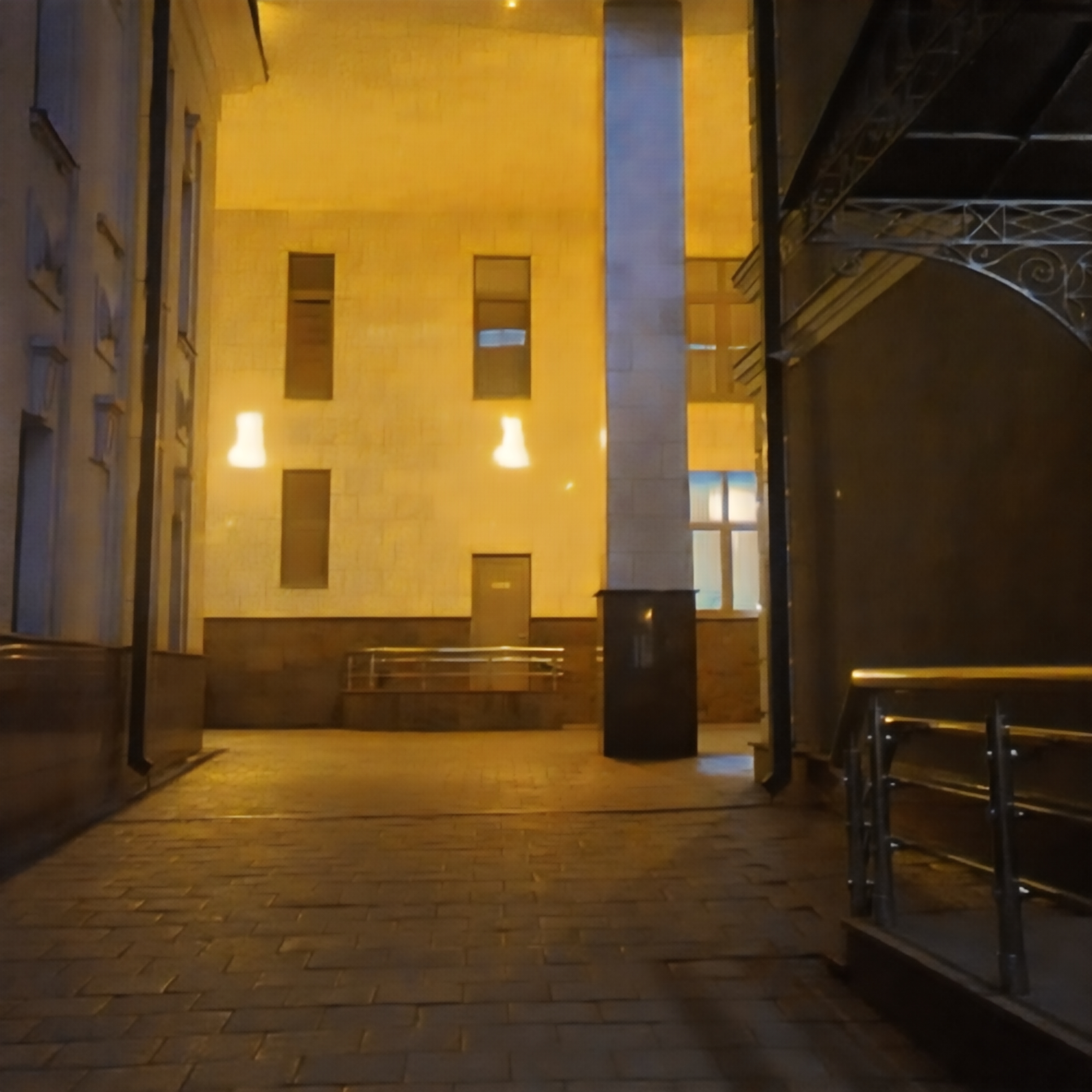}}
  \centerline{sorange}\medskip
\end{minipage}
\hfill
\begin{minipage}[t]{0.16\linewidth}
  \centering
  \centerline{\includegraphics[width=2.82cm]{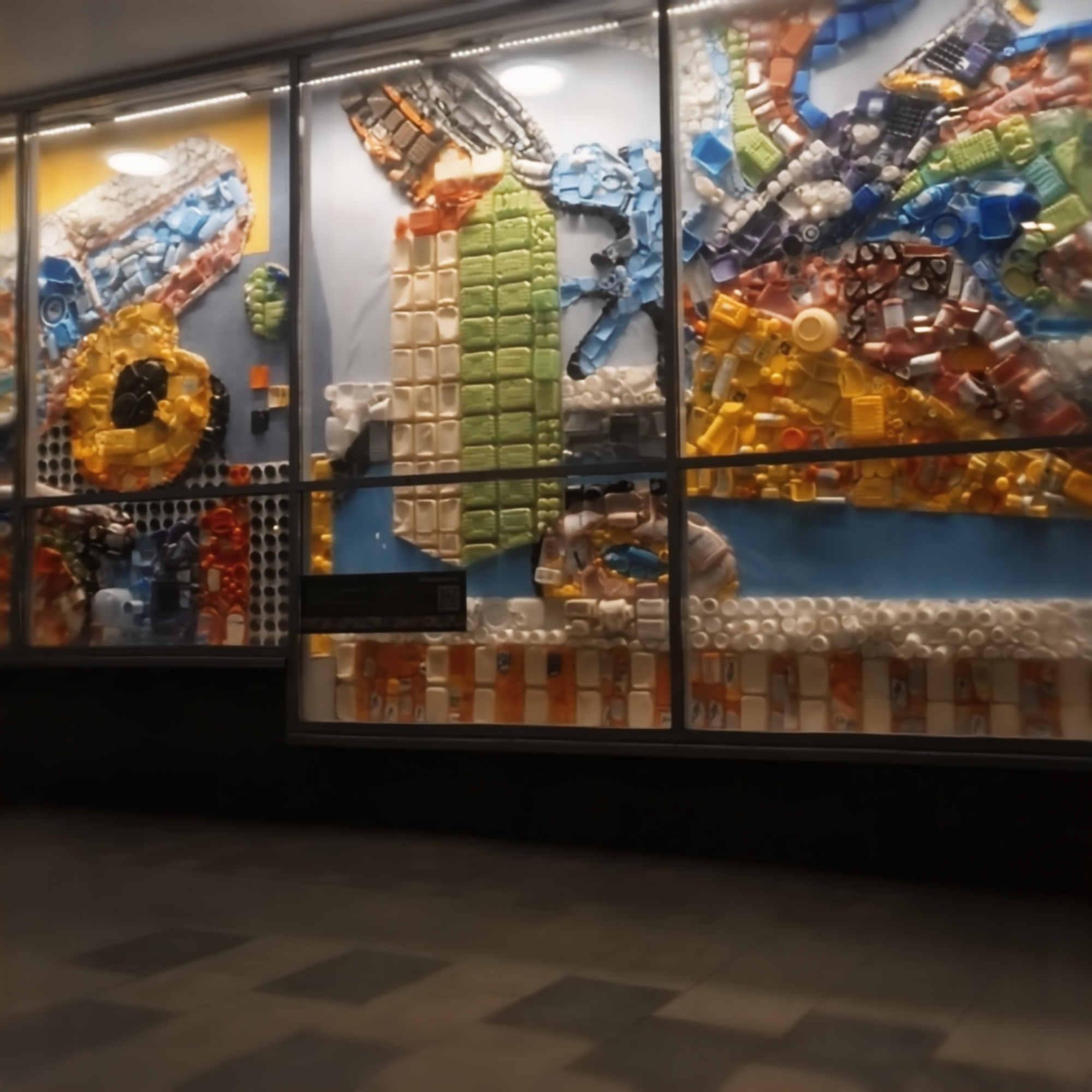}}
  \vspace{0.1cm}
  \centerline{\includegraphics[width=2.82cm]{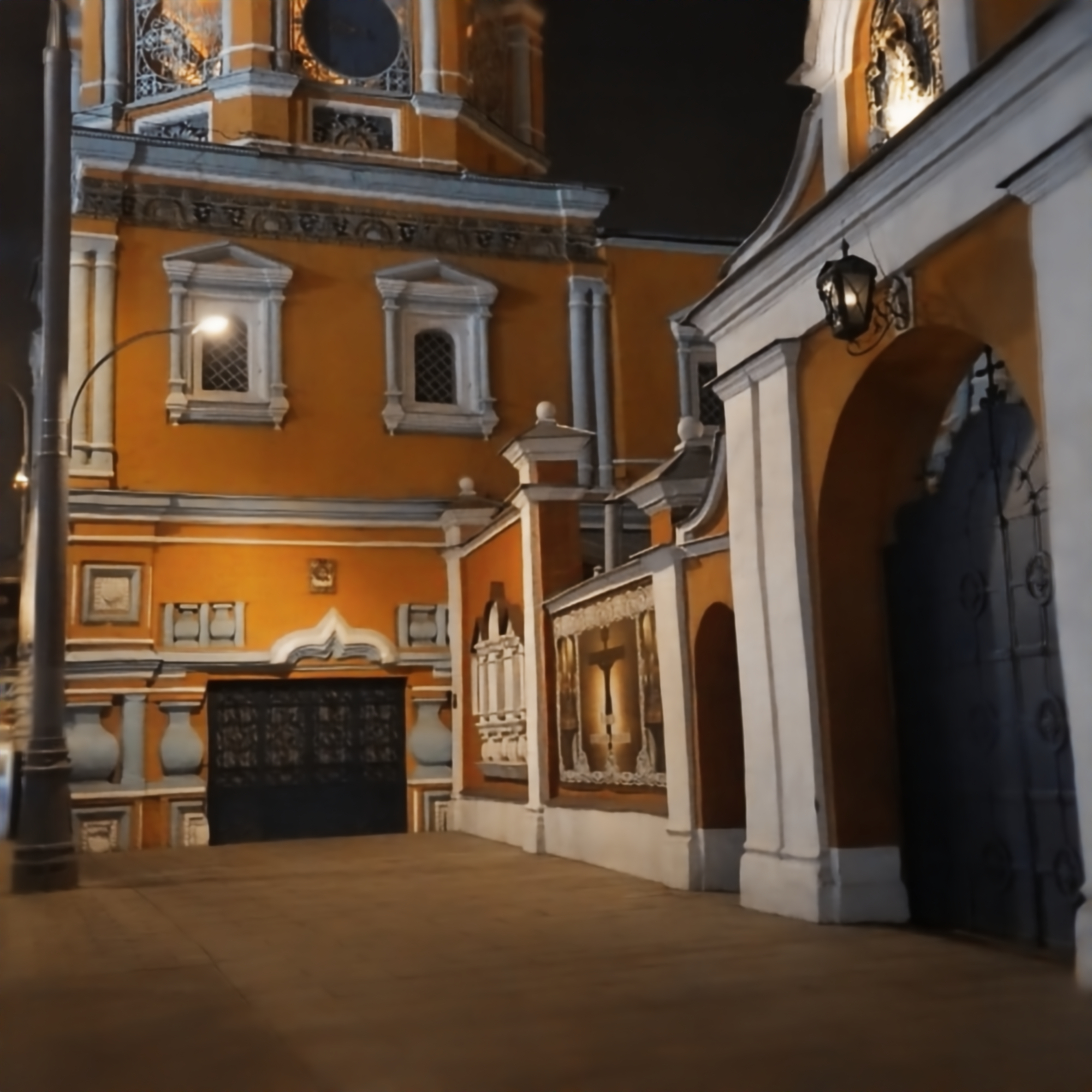}}
  \vspace{0.1cm}
  \centerline{\includegraphics[width=2.82cm]{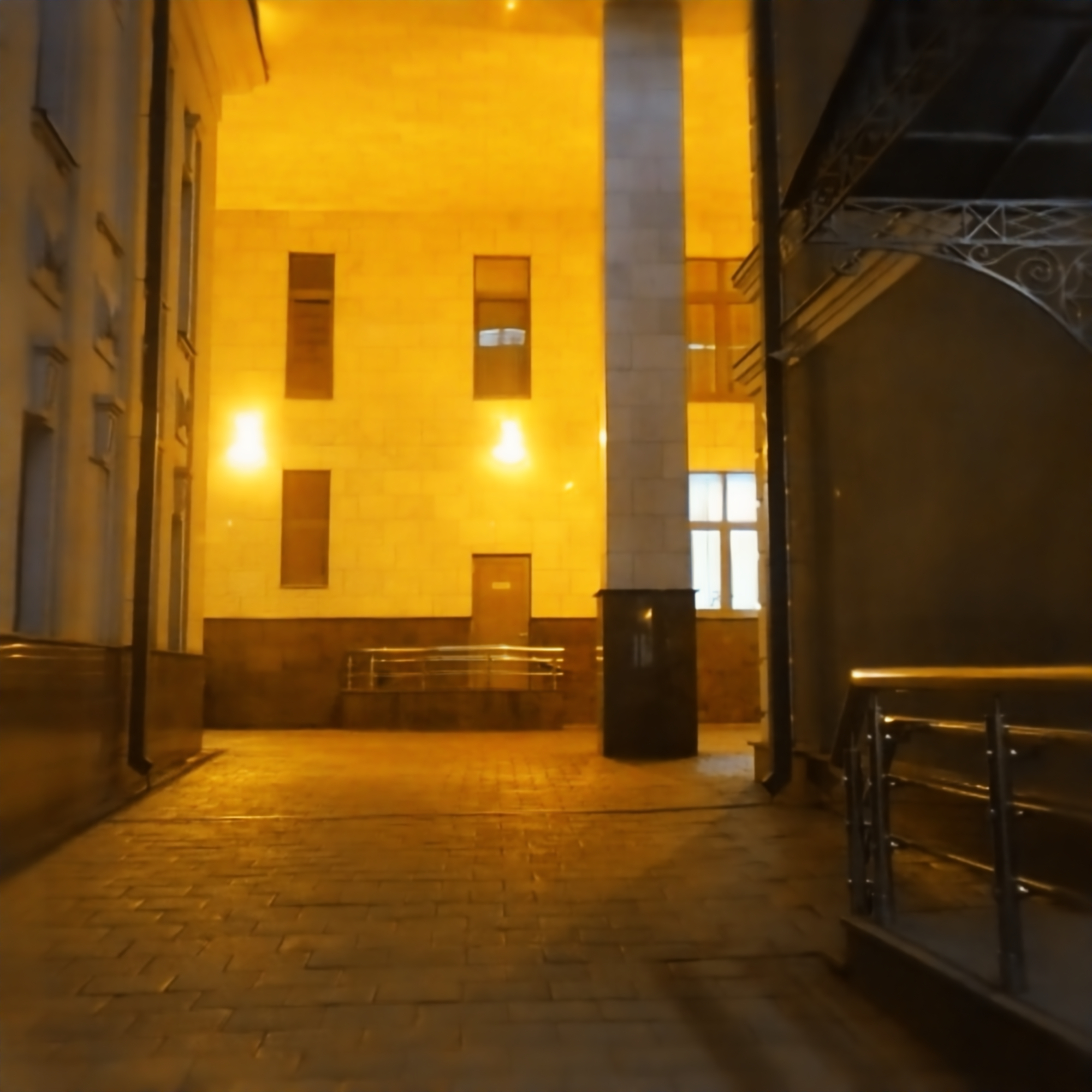}}
  \centerline{NJUST\_KMG}\medskip
\end{minipage}
\hfill
\begin{minipage}[t]{0.16\linewidth}
  \centering
  \centerline{\includegraphics[width=2.82cm]{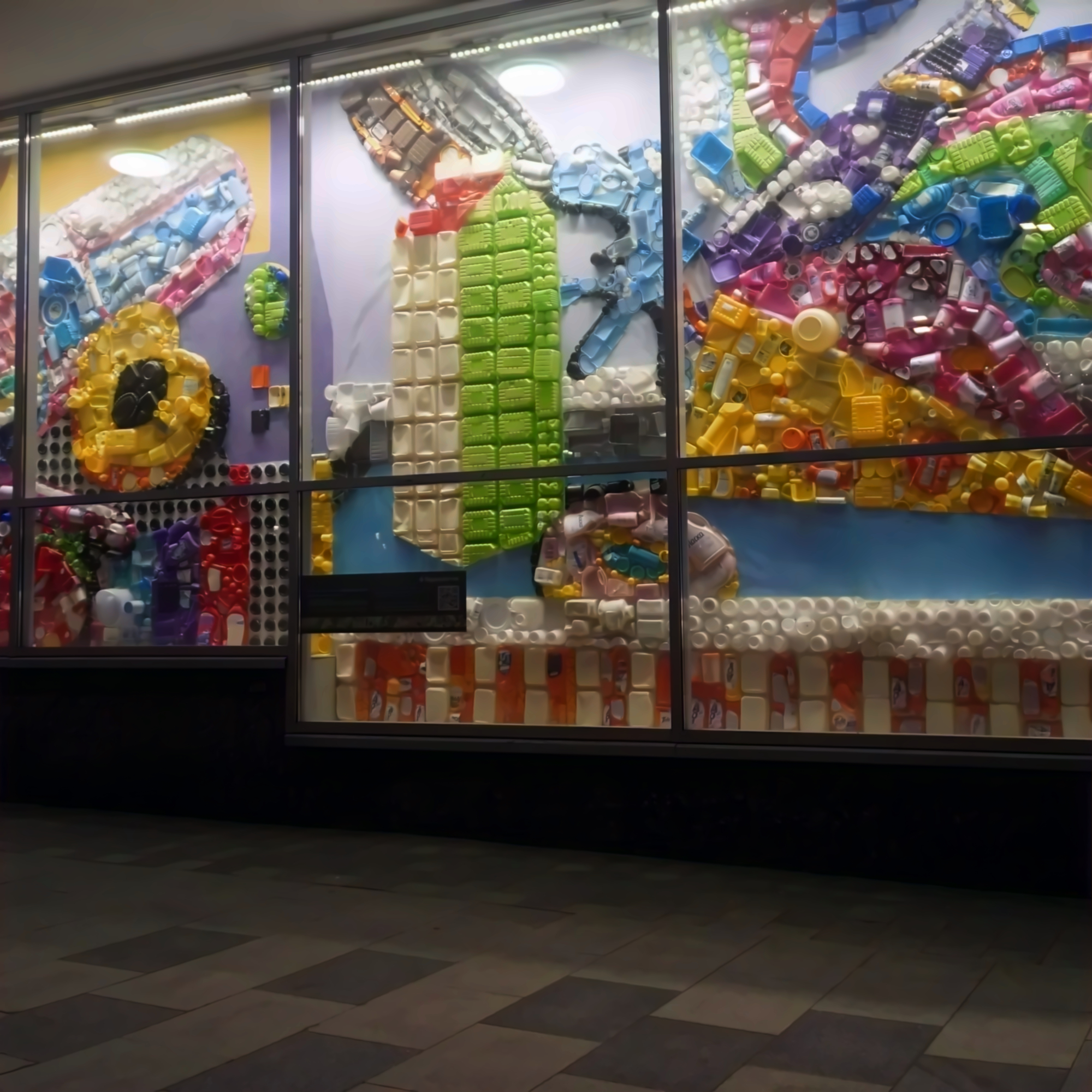}}
  \vspace{0.1cm}
  \centerline{\includegraphics[width=2.82cm]{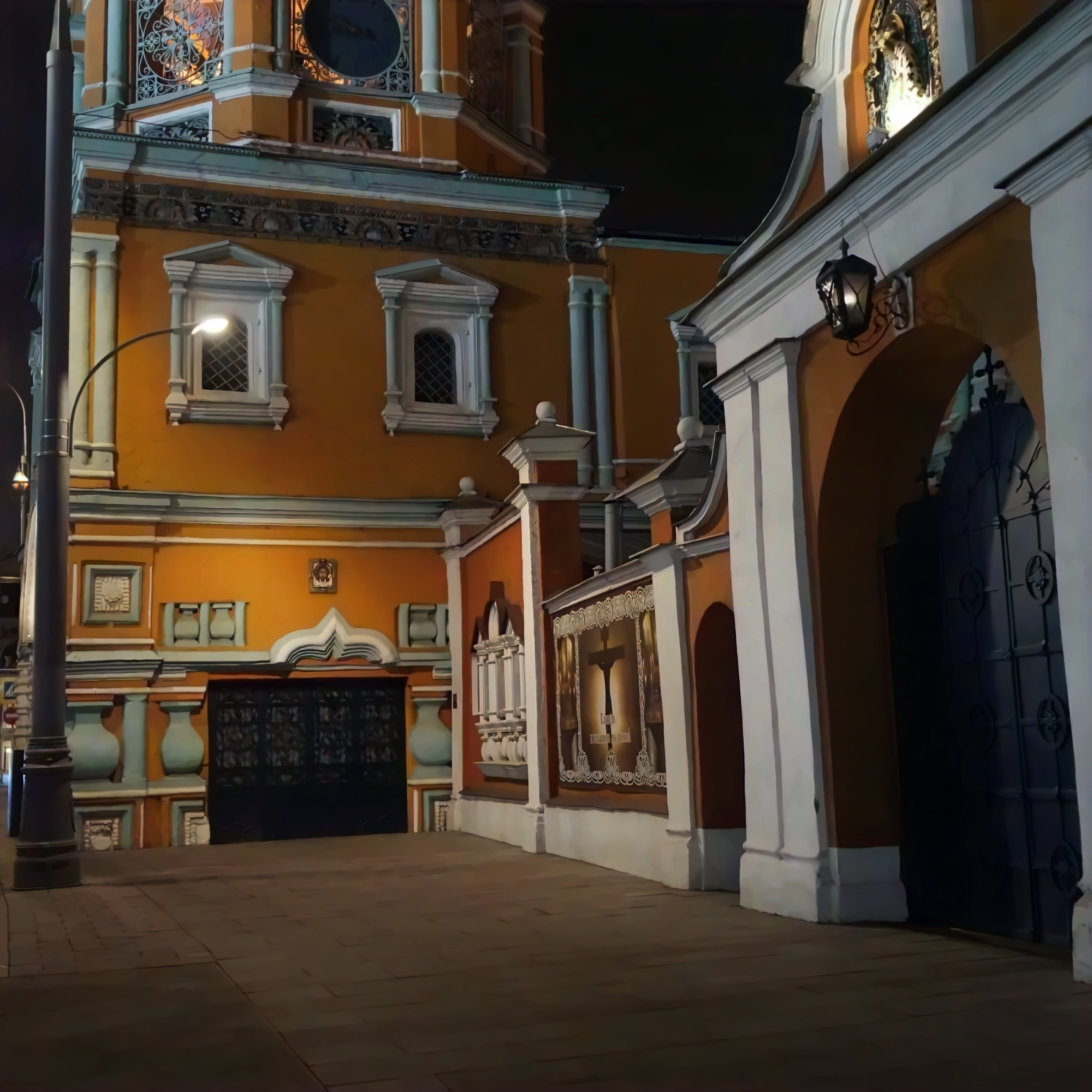}}
  \vspace{0.1cm}
  \centerline{\includegraphics[width=2.82cm]{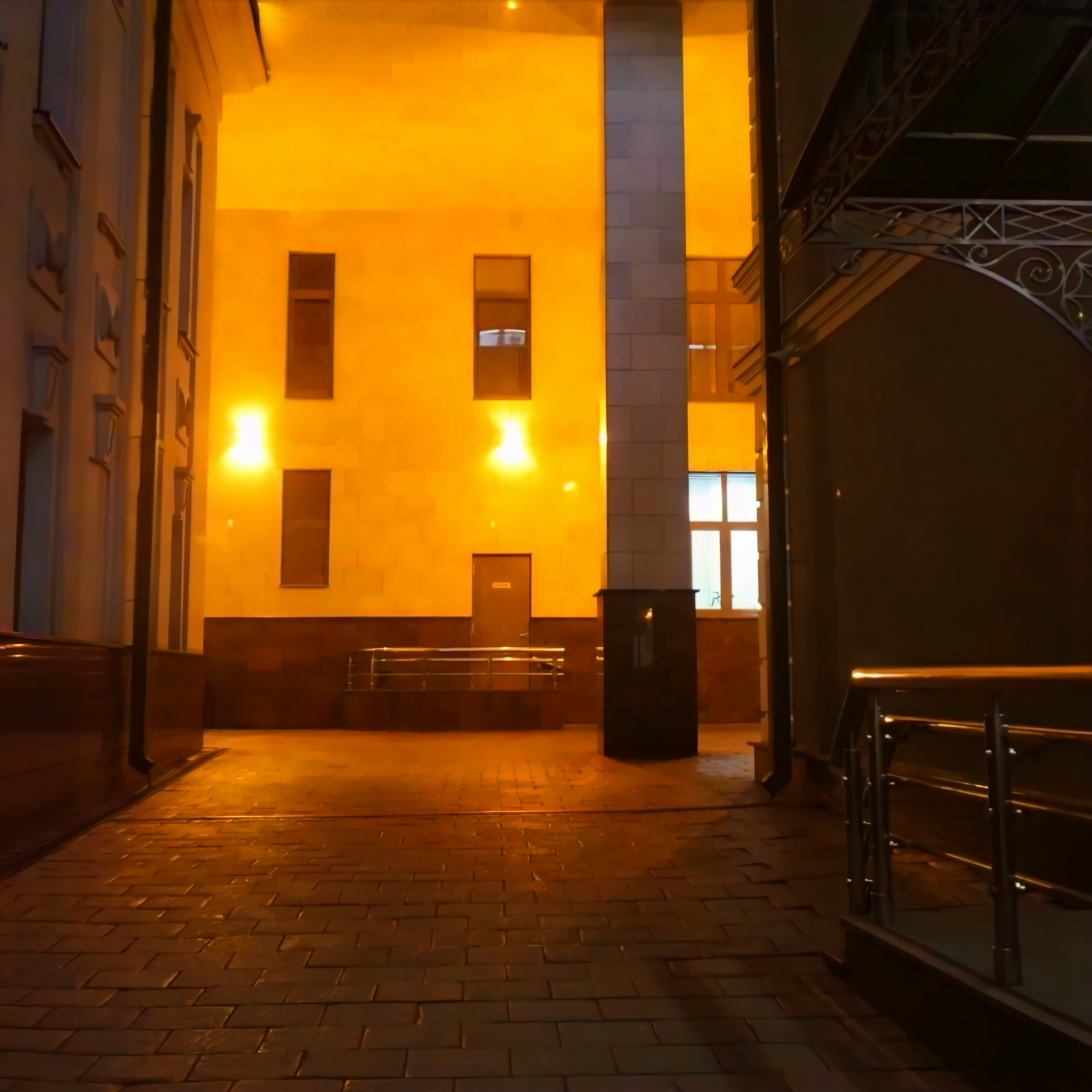}}
  \centerline{Mialgo}\medskip
\end{minipage}
\hfill
\begin{minipage}[t]{0.16\linewidth}
  \centering
  \centerline{\includegraphics[width=2.82cm]{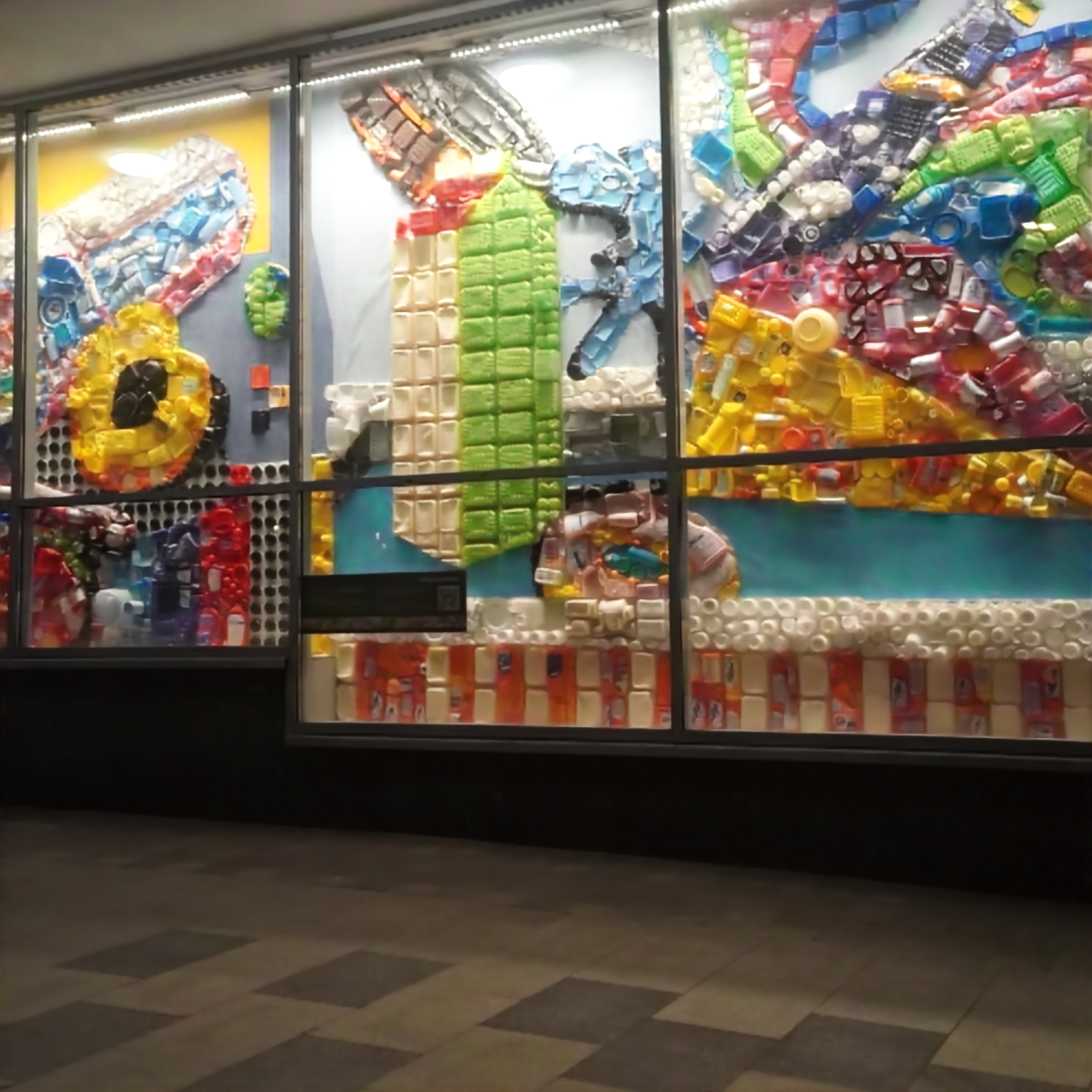}}
  \vspace{0.1cm}
  \centerline{\includegraphics[width=2.82cm]{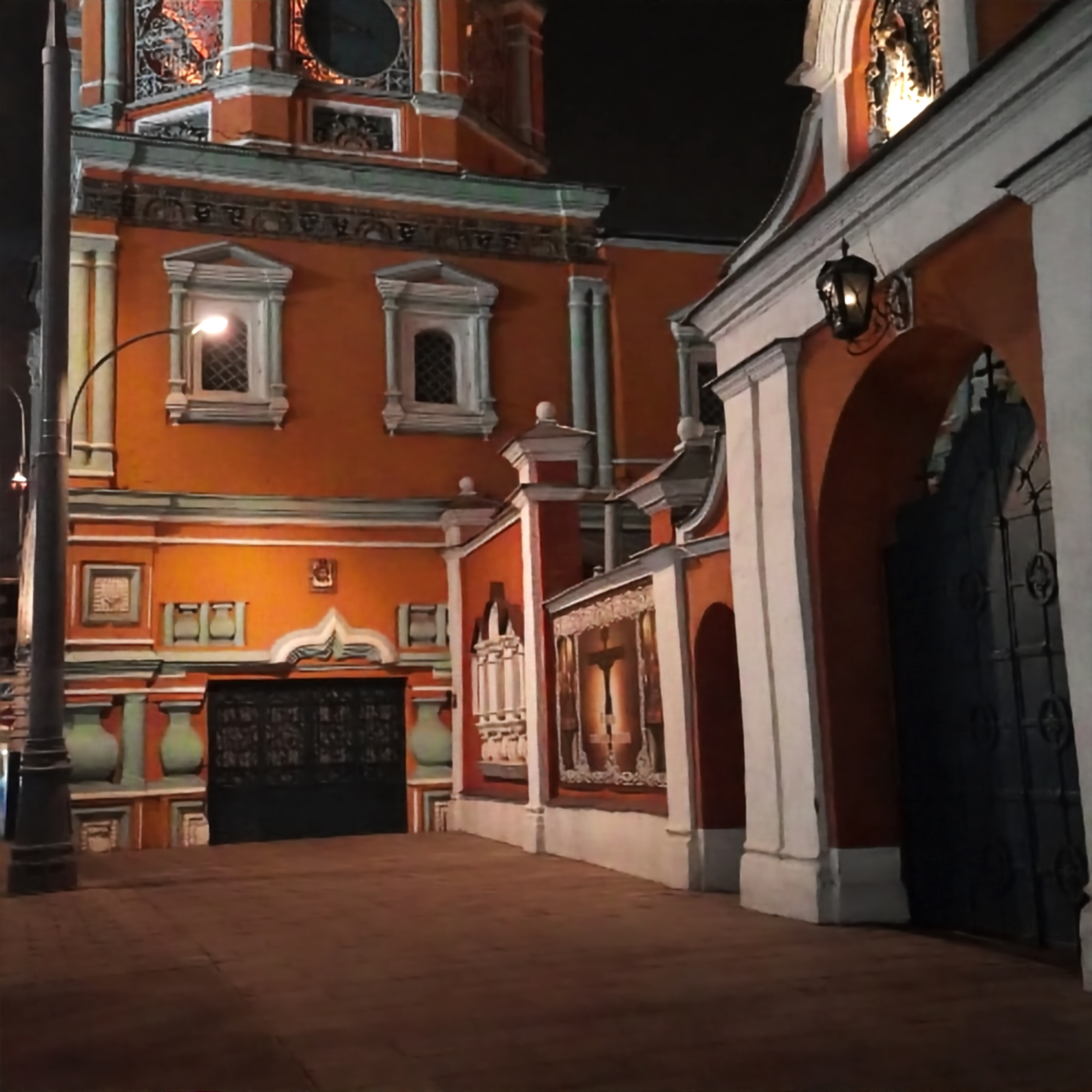}}
  \vspace{0.1cm}
  \centerline{\includegraphics[width=2.82cm]{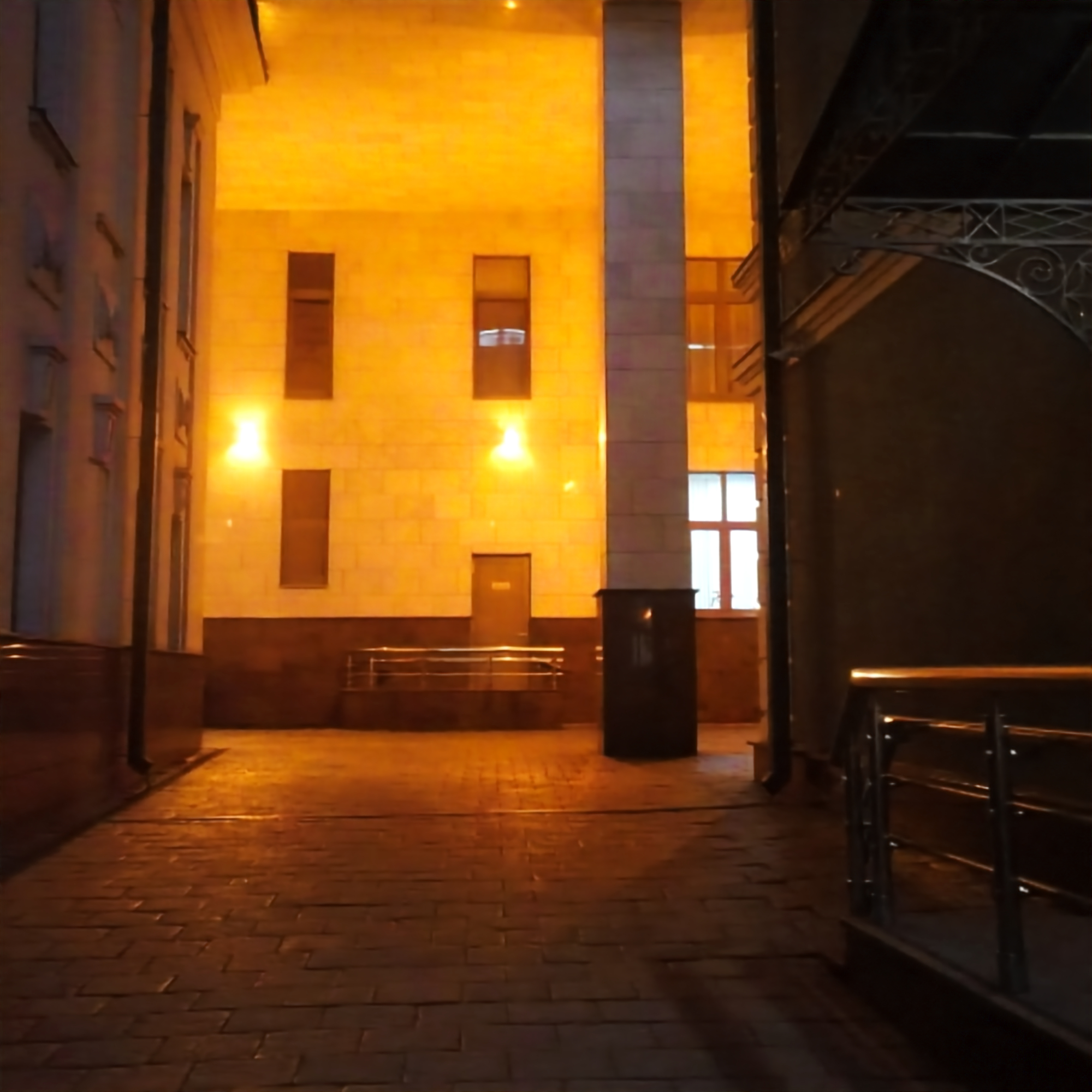}}
  \centerline{Ours}\medskip
\end{minipage}
\hfill
\begin{minipage}[t]{0.16\linewidth}
  \centering
  \centerline{\includegraphics[width=2.82cm]{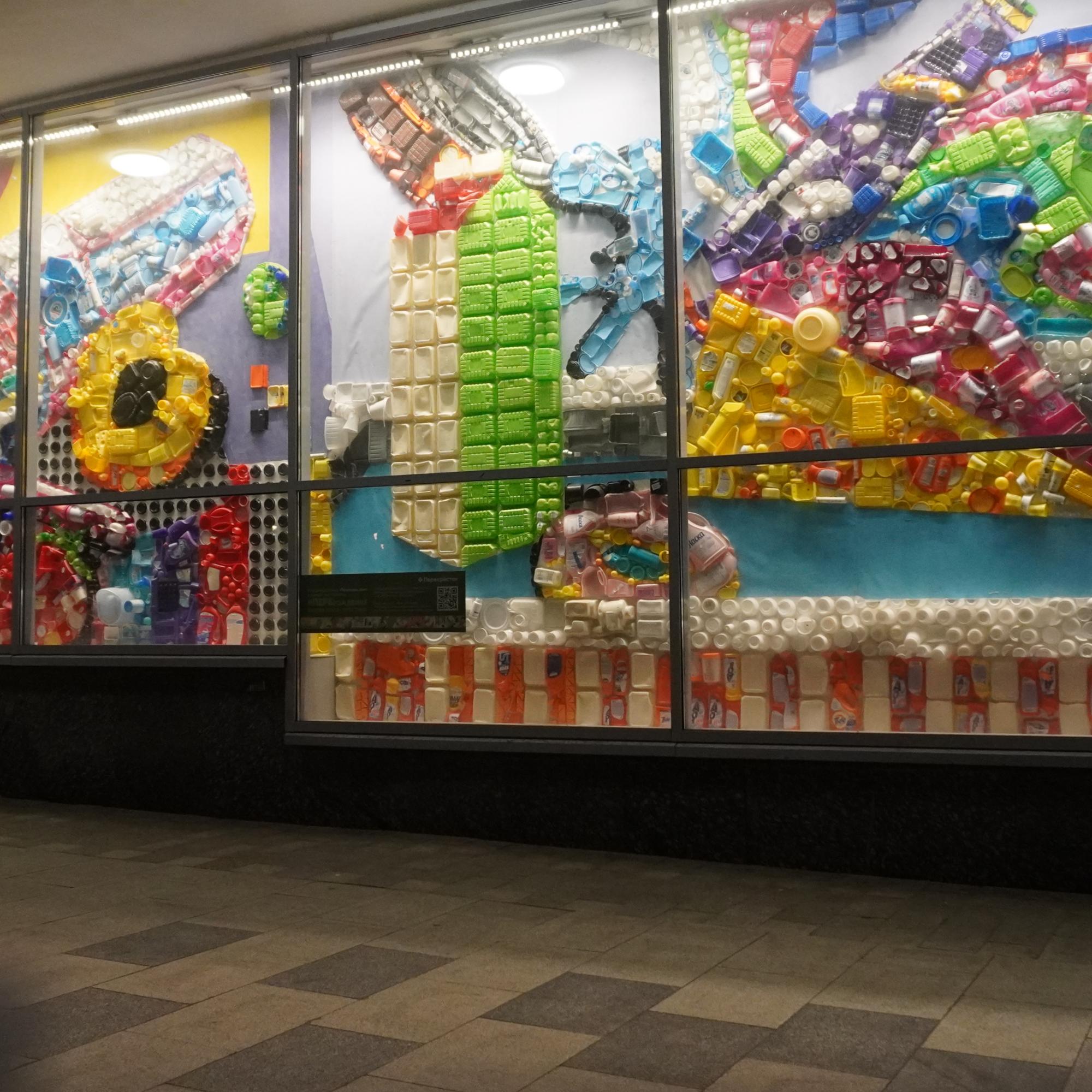}}
  \vspace{0.1cm}
  \centerline{\includegraphics[width=2.82cm]{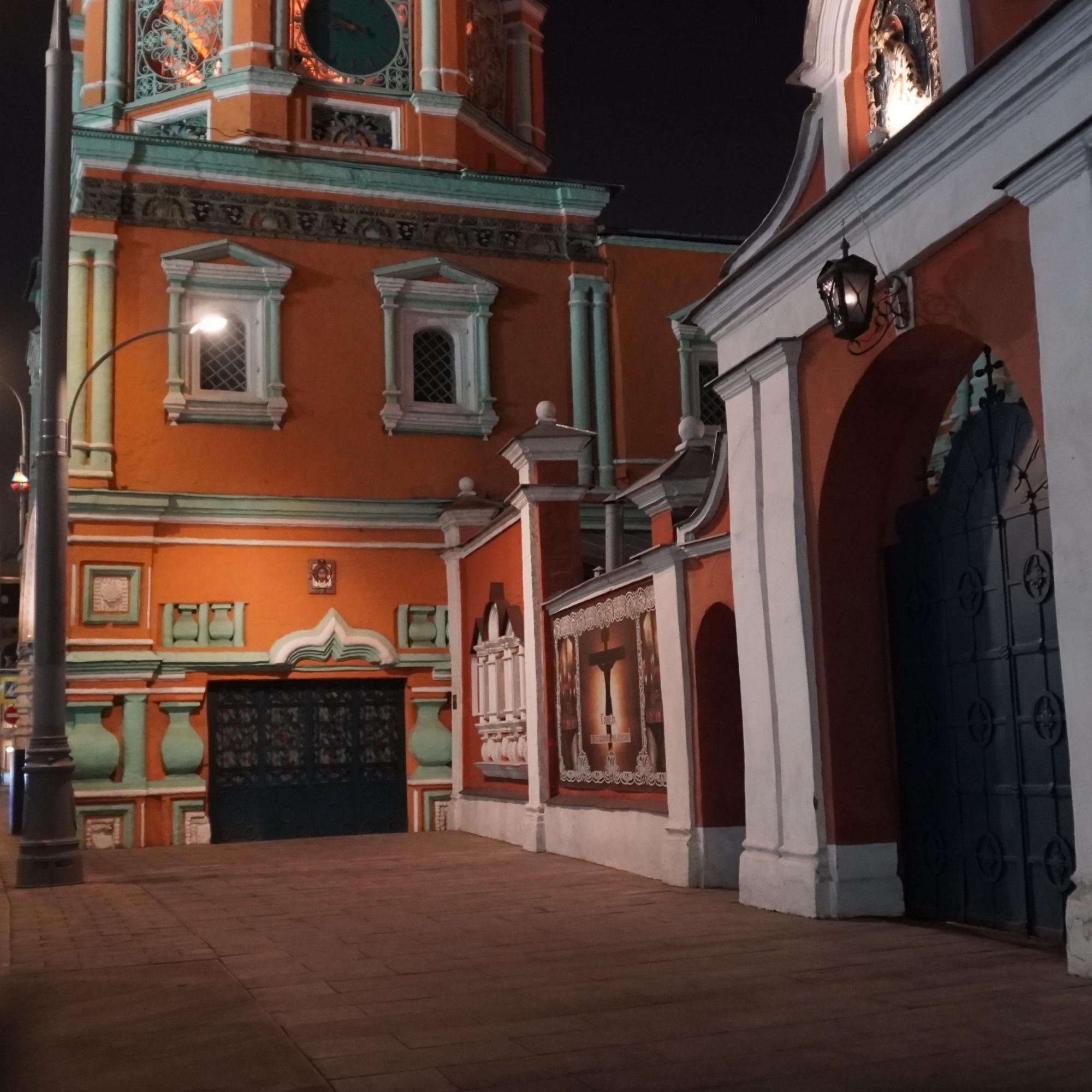}}
  \vspace{0.1cm}
  \centerline{\includegraphics[width=2.82cm]{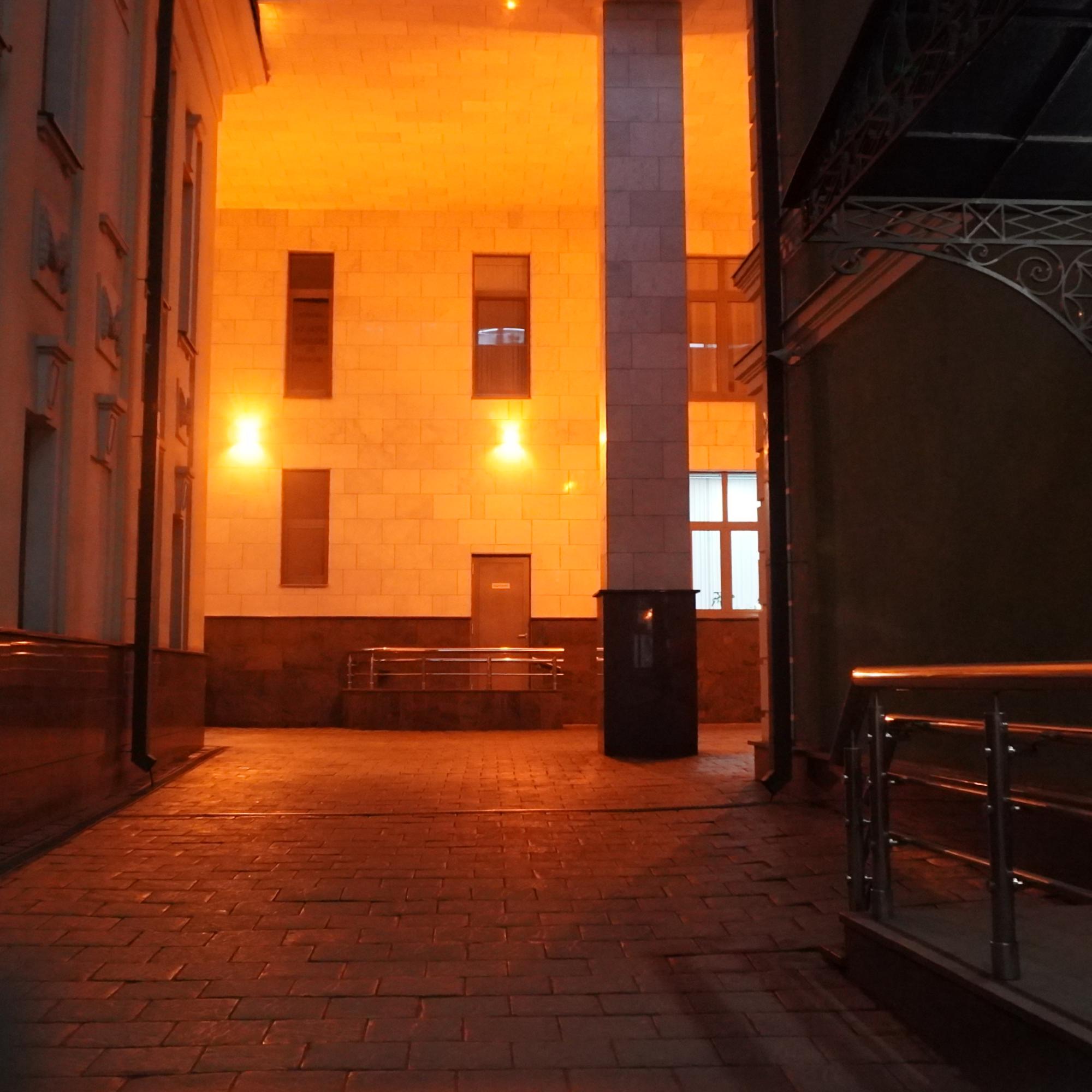}}
  \centerline{GT}\medskip
\end{minipage}
\caption{Qualitative comparison on challenging Night Photography Rendering scenes.}
\label{fig:qual}
\end{figure*}

The substantial improvement is testament to the effectiveness of our integrated design. This gain can be attributed to the strong HVI color space foundation, learning strategy directly derived from the RAW-domain, and the specialized loss function design, which together reduce multi-illuminant color bias and provide output that is both structurally sound and perceptually accurate.

From a qualitative perspective, the superior performance in perceptual metrics directly translates into improved visual quality, as illustrated in Figure~\ref{fig:qual}. Conventional high-fidelity solutions often fail under extreme conditions of NPR due to poor handling of high-dynamic range and multi-illuminant conditions. This causes systematic color inconsistency and a pronounced loss of fine details relative to the ground truth. The core challenge of color constancy at extreme NPR scenes is addressed by the HVI color space foundation and the EFDM-based loss, which collectively provide color fidelity across the image. Simultaneously, the combination of RAW-domain feature processing and Wavelet-based feature propagation ensures high-quality detail preservation, effectively mitigating the noise-detail trade-off inherent in conventional methods. 

\subsection{Ablation Study}

We perform an incremental ablation study, starting from the base HVI-CIDNet~\cite{yan2025hvi}, to rigorously evaluate the additive contribution of each of our four specialized components. The results, summarized in Table~\ref{tab:ablation_study}, demonstrate the need to sequentially integrate these refinements to achieve the final performance gain.

The largest initial performance leap occurs with the addition of the \textit{RGGB-to-Features}, which confirms the vital role of learned RAW-domain demosaicking and noise handling in crafting high-quality input features. The subsequent addition of Wavelet-based feature propagation provides further gains in fidelity and perceptual scores, thus validating its ability to mitigate detail loss during multi-scale processing. Moreover, the specialized loss terms confirm their targeted impact on perceptual metrics where the inclusion of $\mathcal{L}_{\text{FDM}}$ achieves the lowest LPIPS score. The final addition of the sample-based loss weighting strategy (\textit{i.e.}, $\boldsymbol{\alpha}$) acts as the ultimate stabilizer and color corrector: while slightly increasing LPIPS, could be considered as a common trade-off, it delivers the best overall PSNR ($23.63$) along with the lowest color difference ($5.42$).

\begin{table}[!t]
    \caption{Impact of contributions in our proposed framework.}
    \centering
    \label{tab:ablation_study}
    \footnotesize
    \begin{tabular}{l|c|c|c|c}
    \hline
    \textbf{Configuration} & \textbf{PSNR} $\uparrow$ & \textbf{SSIM} $\uparrow$ & $\mathbf{\Delta E} \downarrow$ & \textbf{LPIPS} $\downarrow$ \\
    \hline
    \textbf{HVI-CIDNet (Base)} & 21.71 & 0.757 & 7.06 & 0.418 \\
    \hline
    + RGGB-to-Features & 23.48 & 0.783 & 5.61 & 0.409 \\
    + Wavelet Feat. Prop. & 23.60 & 0.783 & 5.51 & 0.399 \\
    + $\mathcal{L}_{\text{FDM}}$ & 23.58 & \textbf{0.785} & 5.53 & \textbf{0.379} \\
    + Dynamic Loss ($\boldsymbol{\alpha}$) & \textbf{23.63} & \textbf{0.785} & \textbf{5.42} & 0.388 \\
    \hline
    \end{tabular}
\end{table}

\section{Conclusion}

This paper addresses the significant perceptual gap in Night Photography Rendering (NPR) by structurally redesigning the RAW-to-RGB pipeline. We introduced \textit{pHVI-ISPNet}, a unified framework leveraging the HVI color space to manage extreme high-dynamic range and multi-illuminant conditions. Our approach integrates novel architectural improvements into the base CIDNet architecture and utilizes specialized supervisory signals during optimization, with ablation confirming the necessity of these components for stable optimization across the wide exposure range and rigorous color constancy. This approach achieves competitive reconstruction fidelity while establishing new state-of-the-art results in perceptual metrics ($\Delta E$ and LPIPS) on the NTIRE 2025 benchmark.


\vfill\pagebreak

\bibliographystyle{IEEEbib}
\bibliography{refs}

\end{document}